\documentclass[a4paper,fleqn]{cas-sc}

\usepackage[authoryear,longnamesfirst]{natbib}
\usepackage{siunitx}
\usepackage[caption=false,font=normalsize,labelfont=sf,textfont=sf]{subfig}

\def\tsc#1{\csdef{#1}{\textsc{\lowercase{#1}}\xspace}}
\tsc{WGM}
\tsc{QE}
\tsc{EP}
\tsc{PMS}
\tsc{BEC}
\tsc{DE}

\begin{document}
\let\WriteBookmarks\relax
\def\floatpagepagefraction{1}
\def\textpagefraction{.001}
\shorttitle{}
\shortauthors{Hongrui Wu et~al.}

\title [mode = title]{Ambient pressure compensation and robust position control of oil-filled electric joint systems for underwater manipulators}                      



\author[1]{{Hongrui Wu}}[
                        orcid=0000-0002-3465-1026]
\cormark[2]


\affiliation[1]{organization={School of Mechanical Engineering and Automation, Harbin Institute of Technology, Shenzhen},
                postcode={518055}, 
                state={Guangdong},
                country={China}}

\author[1]{{Songhui Wang}}
\cormark[2]

\author[1]{{Xin Wang}}[orcid = 0000-0001-7543-6288
   ]
\cormark[1]





\cortext[cor1]{Corresponding author. School of Mechanical Engineering and Automation, Harbin Institute of Technology, Shenzhen, 518055, Guangdong, China. \it{Email addresses}: wangxinsz@hit.edu.cn}
\cortext[cor2]{The authors made the same contributions.}


\begin{abstract}
Electric joint systems are significant elements of an underwater manipulator for its actuation, drive, and control. Working in an underwater environment, the joints suffer huge ambient pressure. To withstand it, the pressure compensation method is usually deployed, whereas the pressurized oil introduces sealing problems as well as parametric uncertainties and unknown disturbances for the dynamic model of the joint. To tackle these issues, this study proposes a design framework for the underwater oil-filled electric joint. The dynamics of the pressure compensation module is analyzed and the structure of the joint is optimized to seal the internal hydraulic oil. An uncertainty dynamic model of the oil-filled joint is established and a robust position controller is designed based on the structured singular value synthesis ($\mu$-synthesis). Experimental results validate the feasibility of the proposed methods.
\end{abstract}



\begin{keywords}
underwater manipulator \sep electric joint system \sep oil-filled pressure compensation \sep robust position control
\end{keywords}

\maketitle

\section{Introduction}
Integrated on autonomous underwater vehicles (AUVs) or remotely operated vehicles (ROVs), underwater manipulators play a pivotal role in most of the subsea interventions \citep{sivvcev2018underwater}. By the actuation methods, they could be chiefly classified into hydraulic manipulators and electric manipulators. Nowadays, researchers have shown an increasing interest in electric ones because of their high control precision and power efficiency. By deploying compatible electric modular joints, they could be further customized for different tasks (\cite{LIN2026133303}). 

The joint system is one of the most crucial parts of the electric underwater manipulator. It is at the core of the manipulator's perception, actuation, drive, and control. Thus, the joint system has received both academic and industrial attention recently. In the early stage, \citet{yoerger1991design} design an underwater manipulator consisting of 3 electric joints. Each joint is equipped with a servo motor and a reduction system comprising cables and pulleys for zero backlash and low friction. To withstand the high underwater ambient pressure and the conductive seawater, the joint is equipped with rotary cartridge seals and filled with mineral oil pressurized by a passive compensation system. For position control and force control, embedded PID controllers of the ready-made servo motor are utilized \citep{di1988development}. Following this pipeline, \citet{xiao2011development} develop an oil-filled modular joint for a 3 degrees of freedom (DoF) electric underwater manipulator. To reduce the viscous loss introduced by the pressurized oil, the inner housing is polished. Lip seals are deployed to prevent the oil from leakage. For a 7 DoF underwater manipulator integrated with an intervention AUV, \citet{ribas2015auv} demonstrate a reconfigurable modular joint UMA. Brushless direct current motors (BLDC) and harmonic drives are selected for torque output. A single hydraulic circuit is deployed for a compact oil-filled pressure resistance design. To implement the dynamic seal, c-shaped lip rings are mounted between two moving components.

In terms of the commercial progress, Reach Robotics has released 4 product lines on electric underwater manipulators \citep{rosette2024wave}. Relevant joint systems are also available. However, they could only work within a depth of 500 m owing to the absence of the pressure compensation system. Seaeye eM1-7 is an electric underwater manipulator developed by SAAB \citep{lagerby2024force}. It's equipped with modular joints and integrated with pressure compensators, ensuring a working depth up to 7000 m. Designed by Nauticus Robotics, the electric Olymic Arm is also equipped with pressure compensators \citep{s24020666}.

Taken together, these studies and products highlight the demand for pressure compensation in terms of underwater electric joints. Working in an underwater environment, the hollow housing of the joint system is vulnerable to the high ambient pressure. To compensate for the external pressure, it's common for the designers to fill the inside of the joint with pressurized hydraulic oil. However, the introduction of viscous oil would cause dynamic perturbations and unknown disturbances for the joint model, worsening further control. For underwater joints, \citet{yoerger1991design} implement closed-loop position control using a PI controller. Experimental results show that their method fails to eliminate the steady-state error. \citet{guohua2006underwater} develop a servo control system using both analog and digital circuits. It employs a dual-loop design of current and speed control combining Bang-Bang and PID control. \citet{xiao2011development} develop a dynamic model of the oil-filled motor and subsequently design a sliding mode position controller. Without a precise plant model, this controller demonstrates excellent robustness. Neglecting the model of the oil-filled joint, \citet{8665279} combine a fuzzy controller with a PI one to improve the control performance. For underwater oil-filled joints, \citet{liao2021prescribed} propose an observer-based robust control method to enhance the dynamic performance of the system. They treat the oil-filled losses as unknown disturbances and the change of motor dynamics caused by core losses as model parametric uncertainties. Afterwards, a preset performance function (PPF) is utilized to constrain both transient and steady-state trajectory tracking errors. An extended state observer (ESO) is used to estimate the unmeasurable joint velocity signals and the system’s uncertainties. Combining the PPF and ESO functions, the researchers construct a non-singular fast terminal sliding mode controller (NFTSMC). The simulation shows that the trajectory tracking error would tend to zero under appropriate parameters. From these studies, it can be seen that most of the control methods neglect the varying model of the oil-filled joint and lack a specific technique to handle it (\cite{yang2023disturbance,yang2025robust}). Meanwhile, relevant experiments are deficient.


This paper aims to develop a general design framework for underwater electric joint systems considering ambient pressure compensation and robust position control. The compensation module balances the huge ambient pressure to protect the joint structure from cracking. Although the resulting pressure difference from the inside out avoids the intrusion of the conductive water, it causes the leakage of the hydraulic oil. Therefore, we also optimize the joint structure for better sealing performance. As mentioned before, the pressurized compensation oil worsens the control. Thus, we design a robust position controller to tackle the introduced dynamic perturbations and unknown disturbances. The main contributions of this study are summarized as follows:

\begin{enumerate}[(1)]
\item The joint system considering dynamic perturbations and unknown disturbances caused by the pressurized compensation oil is modeled and controlled based on the $\mu$-synthesis.
\item The pressure difference caused by the compensation module is analyzed. A joint structure optimization method is designed to avoid the leakage under this pressure difference.
\item A mechatronic design framework for underwater electric joint systems is proposed including mechanical, drive and control modules. 
\end{enumerate}
The remainder of this paper is arranged as follows: Section \ref{sec2} first designs the mechanical module and the drive module of the joint system. To withstand the ambient pressure, Section \ref{sec3} analyzes the dynamics of the pressure compensation system and optimally designs the joint structure for better sealing performance. Considering the effect of the pressurized compensation oil, Section \ref{sec4} establishes an uncertainty model of the joint system and develops a robust controller to handle the dynamic perturbations and unknown disturbances using $\mu$-synthesis. Section \ref{sec5} conducts experiments to validate the proposed control framework. Finally, Section \ref{sec6} reaches conclusions of this study.

\section{Mechatronic module integration of the underwater joint system}\label{sec2}
For the introduction of the research subject, we start with the integration of the underwater joint's mechatronic module expected to work at a depth of 1000 m. Basically, the joint system consists of a mechanical module, a drive module, and a control module. The control module receives commands from the upper-level planning module. Based on the feedback of the joint, it implements a control algorithm and outputs direct current (DC) signals to the drive module. Subsequently, the drive module converts the DC control signals into amplified alternating current (AC) signals to operate the mechanical module. Apart from realizing the basic functions of a joint system, it's necessary to consider the component selection, the pressure resistance, and the sealing techniques due to the underwater environment, where the ambient pressure is rather huge. To address it, an oil-filled compensation system is deployed. More details on pressure compensation and sealing will be further discussed in Section \ref{sec3}.

\begin{figure}[!h]
   \centering
   \includegraphics[width=.7\textwidth]{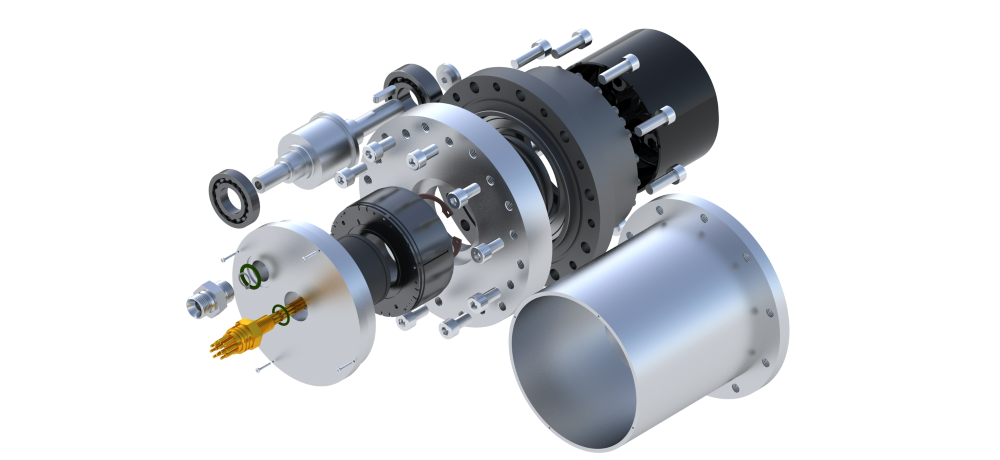}
   \caption{Mechanical module of the joint.}
   \label{FIG:1}
\end{figure}

To output torque and feed back position, the mechanical module is integrated with a frameless torque motor, a harmonic drive, and a resolver. The frameless torque motor is a type of permanent magnet synchronous motor (PMSM). It efficiently and reliably delivers mechanical energy to the harmonic drive for a lower speed and higher torque. Owing to the optoelectronic components, the most common position sensors --- encoders, are highly sensitive and vulnerable to the underwater environment. In contrasts, composed of iron cores and coils, resolvers ensure resistance to the harsh underwater environment of low temperature and high pressure. Further, the compact structure can also endure high-speed rotations and heavy loads under extreme working conditions. Thus, resolvers are selected for underwater position sensing. After the selection of the components, housings and shafts are integrated to accommodate them. To provide necessary interfaces for the joint system in the underwater environment, an underwater electrical connector and a hydraulic adapter are mounted on the end cap of the housing. Fig. \ref{FIG:1} shows the mechanical module of the joint.

The drive module adopts Space Vector Pulse Width Modulation (SVPWM) architecture, which uses a three-phase inverter to convert DC bus voltage into three-phase AC ones. Therefore, a rotating magnetic field could be generated within the stator to rotate the permanent magnet rotor based on a common DC power supply. To decouple and simplify the control, Clark and Park transformations are applied to  convert the three-phase AC signals into two-phase DC signals.

To realize SVPWM, a microcontroller is deployed to implement motor drive algorithms. It outputs Pulse Width Modulation (PWM) signals from timer peripherals to a gate driver, where the PWM signals can be amplified to high-current inputs to open the gates of the transistors in the three-phase inverter. Besides, shunt resistors are used to sample the three-phase AC signals on the motor's windings. Via an analog-to-digital converter (ADC), the sampled value is sent to the microcontroller for further processing. Apart from these common drive components, a resolver-to-digital converter is necessary for our drive module. Despite the advantages mentioned before, the processing of resolver signals is demanding. For the input, resolvers need a sinusoidal excitation signal. For the output, resolvers could only transmit AC voltage signals whose magnitudes are related to the motor position. To handle the issues, the resolver-to-digital converter is deployed to provide the excitation input for the resolver and calculate the position based on the AC signals for the driving and control algorithms.

\begin{figure}[!h]
   \centering
   \includegraphics[width=.9\textwidth]{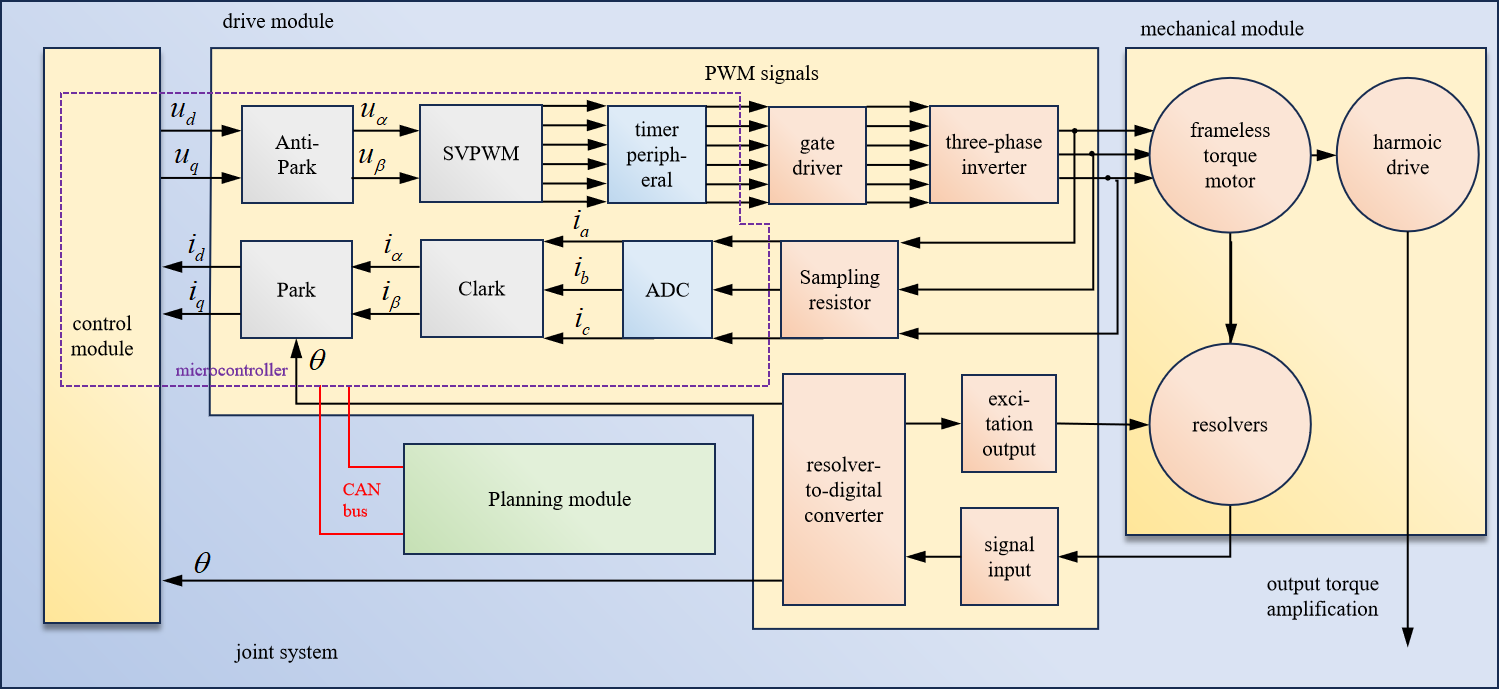}
   \caption{Framework of the mechatronic module.}
   \label{FIG:2}
\end{figure}

In terms of the control module, the relevant algorithms are also implemented by the microcontroller in the drive module. Details of the control algorithm are discussed in Section \ref{sec4}. Fig. \ref{FIG:2} shows the framework of the mechatronic module.

\section{Ambient pressure compensation and sealing optimization}\label{sec3}
Unlike common robotic arms that work on land, the electric joint system in this study operates in an underwater environment at a depth of 1000 m where the ambient pressure is rather huge. Owing to the cavity of the mechanical structure, the housing of the joint can easily break under the pressure. Meanwhile, because of the difference between the external and internal pressures of the housing, conductive liquids in the environment can permeate into the joints, resulting in unacceptable short circuits. To handle these issues, the pressure resistance and sealing techniques are emphasized in this study.

Typically, there are two methods for underwater devices to withstand the ambient pressure: direct pressure resistance and pressure compensation. Being free of additional equipments, direct pressure resistance utilizes material with sufficient strength and stiffness to manufacture the housing of the joint. However, the density of the qualified material that can resist the huge underwater pressure is usually high, leading to a low payload-weight ratio for the underwater manipulator. Not only does the bulky structure worsen the performance of the robotic arm, it also complicates further driving and control.

In contrast, pressure compensation method fills the joint with hydraulic fluids and deploys a pressure compensator to balance the internal oil pressure with the external ambient one. Thus, the net pressure acting on the joint housing is close to 0. When the ambient pressure changes due to depth variation, the elastic material of the pressure compensator deforms, modulating the volume of the hydraulic fluids. Hence, the external pressure is transmitted to the internal oil until the pressure reaches equilibrium. 

\begin{figure}[!h]
\centering
\subfloat[]{\includegraphics[width=.4\textwidth]{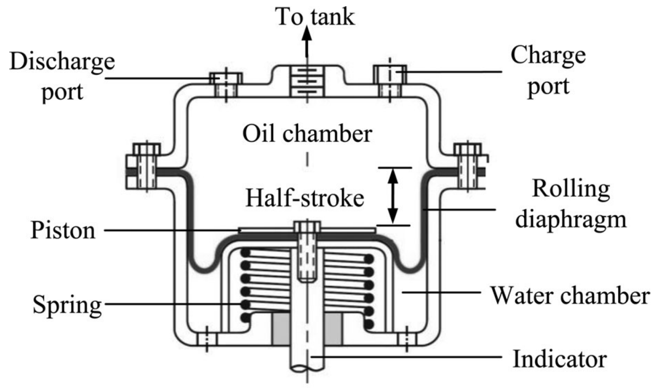}%
\label{fig_a}}
\hfil
\subfloat[]{\includegraphics[width=.3\textwidth]{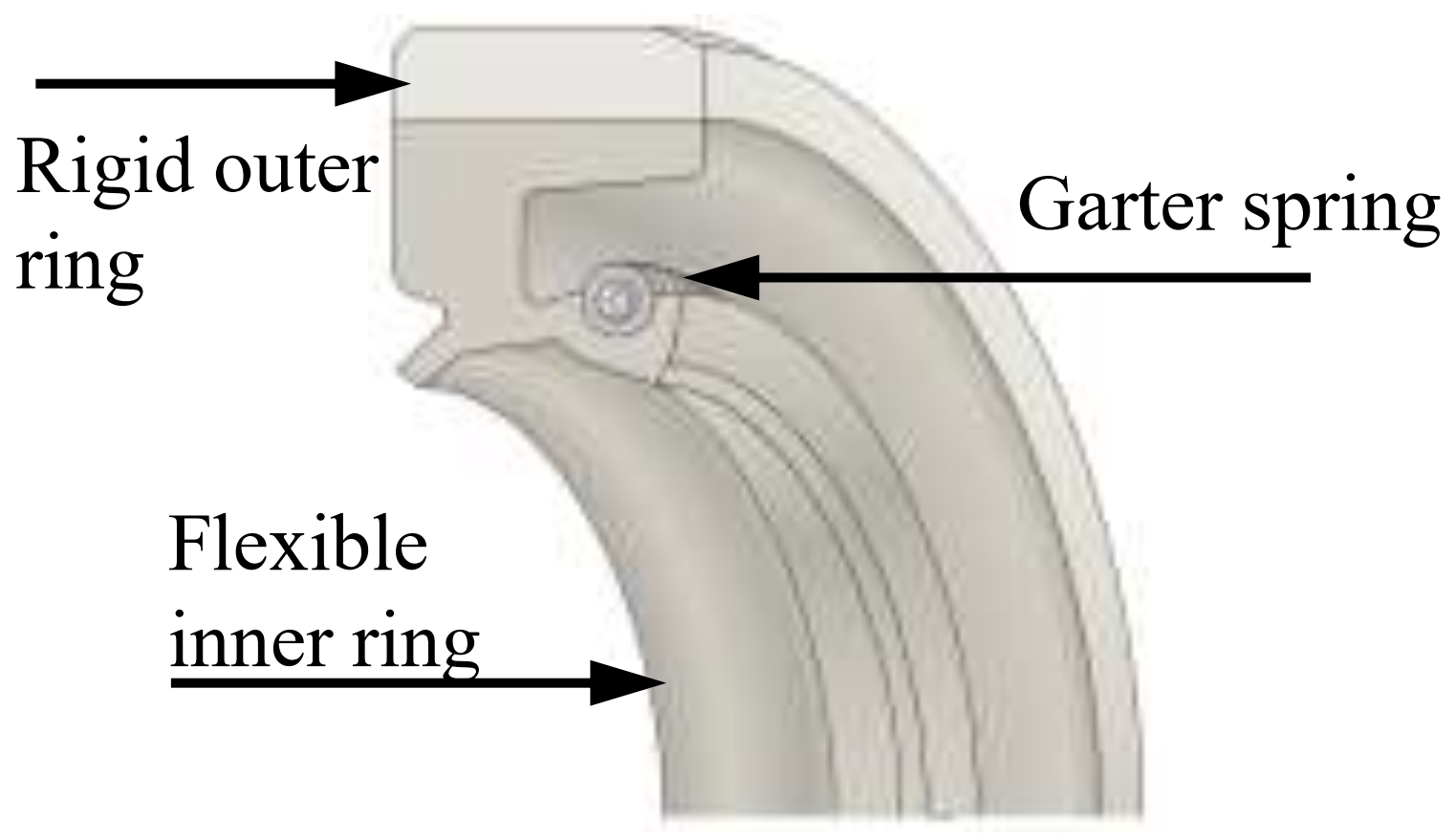}%
\label{fig_b}}
\hfil
\subfloat[]{\includegraphics[width=.17\textwidth]{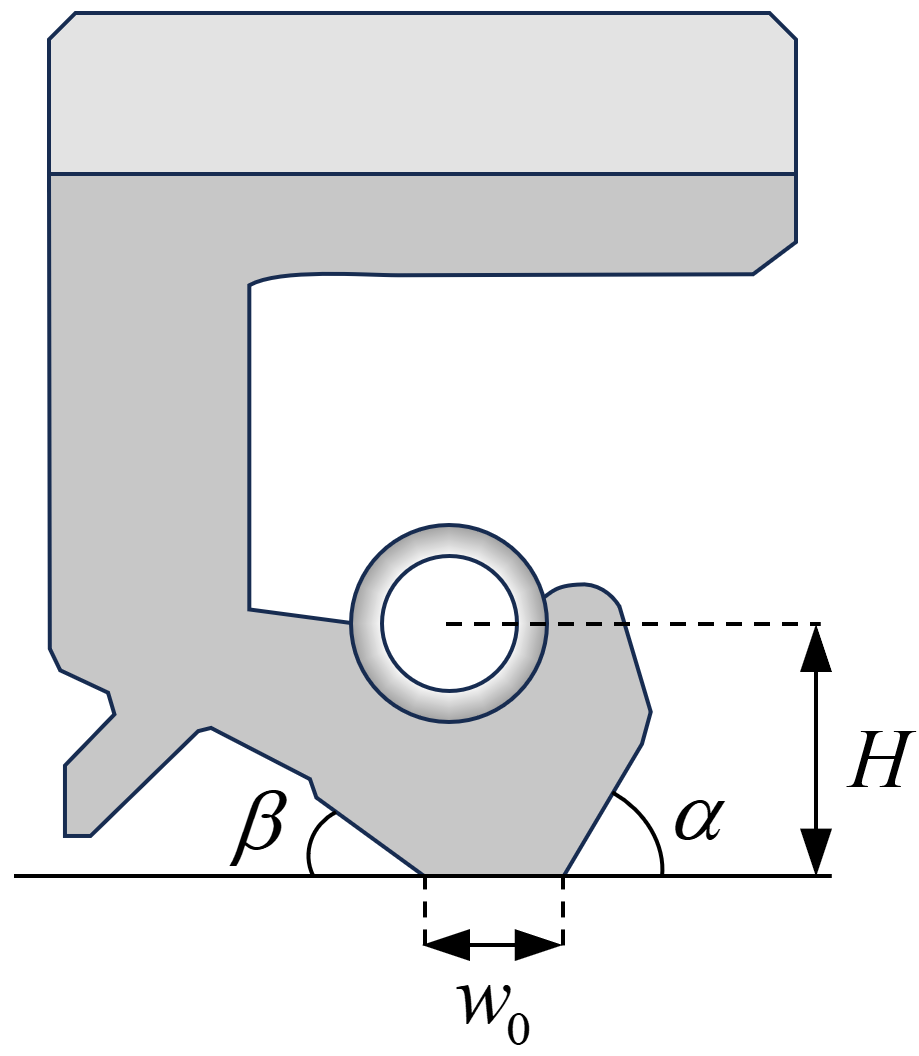}%
\label{fig_c}}
\caption{Schematics of the pressure compensation and the sealing equipments. (a) Pressure compensator. (b) Rotary lip seal. (c) Lip structure.}
\label{fig_schematic}
\end{figure}

Pressure compensators are classified based on the type of the elastic material. In this study, the rolling diaphragm compensator is selected, which enjoys the advantages of large deformation in both directions, lower friction, quicker response and good sealing performance. Divided into a rod chamber and a non-rod chamber, the pressure compensator consists of a rolling diaphragm, a piston and a spring. The spring is installed on the piston rod. Separated by the rolling diaphragm, the rod chamber is connected to the underwater environment, and the oil-filled non-rod chamber is connected to the joint system through a hydraulic pipe. Fig. \ref{fig_schematic}-(a) shows the schematic of the pressure compensator.

For the hydraulic fluids in the pressure compensation system, silicone oil is selected considering its insulation and low viscosity. It avoids short circuits and decreases motor losses due to the friction between the oil and the motor. The dynamic viscosity is a key factor when calculating the viscous torque imposed on the oil-filled joint. In the underwater environment, it significantly ascends with the increase of the depth and the decrease of the temperature. According to Barus equation, this effect could be expressed as:
\begin{equation}
\mu_1 = \mu_{0}e^{\frac{p_1}{A(T_1)+B(T_1)p_1}}
\end{equation}
where $\mu_0$ is the rated dynamic viscosity, $A$ and $B$ are coefficients related to temperature and, $\mu_1$ is the dynamic viscosity under a pressure of $p_1$ and temperature of $T_1$. A silicone oil with a rated kinematic viscosity of 50 cSt is selected as the hydraulic medium of the oil-filled joint. At a depth of 1000 m, the ambient temperature is near 2$^{\circ}$C. Table \ref{tbl1} shows the relevant viscous parameters of this oil \citep{cai2016effect}.

\begin{table}[width=.9\linewidth,cols=4,pos=h]
\caption{Viscous parameters of the silicone oil under 1000 m and 2$^{\circ}$C.}\label{tbl1}
\begin{tabular*}{\tblwidth}{@{} LLLL@{} }
\toprule
Parameter & Definition & Value & Unit\\
\midrule
$\upsilon_0$ & rated kinematic viscosity & \num{50} & cSt \\
$\mu_0$ & rated dynamic viscosity & \num{71.3e-4} & $\text{Pa}\cdot\text{s}$ \\
$A$ & coefficient related to temperature & \num{9.43e7} & - \\
$B$ & coefficient related to temperature & \num{2.8e-1} & - \\
\bottomrule
\end{tabular*}
\end{table}

When the pressure compensator is in the steady-state, the internal pressure of the joint is slightly higher than the external ambient pressure due to the elastic force of the spring on the piston rod \citep{wang2014design}:
\begin{equation}\label{eq2}
    {{p}_{j}}={{p}_{w}}+{{p}_{k}}
\end{equation}
where ${p}_{j}$, ${p}_{w}$, and ${p}_{k}$ are the pressures of the oil, the environment, and the spring, respectively. Ignoring the gravity of the piston and the deformation resistance of the rolling diaphragm, Eq (\ref{eq2}) indicates that the slight pressure difference ${p}_{k}$ protects the joint from external water permeation. However, it would cause the leakage of the internal silicone oil, worsening the performance of the pressure compensation system. 

To handle the pressure difference from the inside to the outside, we calculate the value of ${p}_{k}$ at a depth of 1000 m and a temperature of 2$^{\circ}$C to guide the further design of sealing system. Before getting into the water, the non-rod chamber of the compensator is precompressed with silicone oil and brings about an initial spring displacement $x_0$, corresponding to the neutral
plane position. With the increase of the depth, the spring stretches. Therefore, the volume of the non-rod chamber and the joint system ${V}_{c}$ becomes:
\begin{equation}
    {{V}_{c}}={{V}_{o}}-{{S}_{r}}x
\end{equation}
where ${V}_{0}$ is the volume in the neutral plane position, ${S}_{r}$ is the effective area of the rolling diaphragm, and $x$ is the displacement of the piston rod from the neutral plane. Along with the pressure increase and temperature decrease, the volume of the silicone oil also changes:
\begin{subequations}
\begin{align}
\Delta {{V}_{T}}&=\alpha {{V}_{c}}\Delta T\\
\Delta {{V}_{p}}&=-\frac{{{V}_{c}}}{K}\Delta {{p}_{j}}
\end{align}
\end{subequations}
where $\Delta {{V}_{T}}$ and $\Delta {{V}_{p}}$ are the volume changes of the silicone oil caused by temperature variation $\Delta T$ and pressure variation $\Delta {{p}_{j}}$, respectively, $\alpha$ is the thermal expansion coefficient of the silicone oil, and $K$ is the bulk modulus of the silicone oil. Based on the analysis before, the continuity equation of the pressure compensation system is expressed as:
\begin{equation}\label{eq5}
Q=\frac{d{{V}_{T}}}{dt}+\frac{d{{V}_{p}}}{dt}-\frac{d{{V}_{c}}}{dt}=\alpha {{V}_{0}}\frac{dT}{dt}-\frac{{{V}_{0}}}{K}\frac{d{{p}_{j}}}{dt}-{{S}_{r}}\frac{dx}{dt}
\end{equation}
where $Q$ is the volumetric flow rate of the system. Considering the slight volume variation, we substitute the constant $V_0$ for the variable ${V}_{c}$ in Eq (\ref{eq5}) for simplification \citep{wang2013dynamic}.

In the meanwhile, the dynamic equation of the pressure compensation system could be expressed as:
\begin{equation}\label{eq6}
{{m}_{c}}\frac{{{d}^{2}}x}{d{{t}^{2}}}={{m}_{c}}g+{{p}_{j}}{{S}_{r}}-{{p}_{w}}{{S}_{r}}-k({{x}_{0}}+x)-c\frac{dx}{dt}
\end{equation}
where $m_c$ is the mass of the piston, $g$ is the standard gravity, and $c$ is the coefficient of viscous friction.

\begin{figure}[!t]
   \centering
   \includegraphics[width=.5\textwidth]{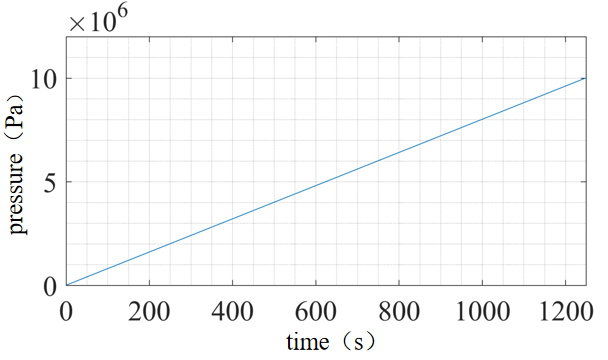}
   \caption{Graph of the joint pressure $p_j$.}
   \label{fig_pj}
\end{figure}

Within the continuity equation Eq (\ref{eq5}) and the dynamic equation Eq (\ref{eq6}), the time variations of the external ambient pressure $p_w$ and the temperature $T$ are expressed based on the actual situation:
\begin{subequations}
\begin{align}
{{p}_{w}}\left( t \right)&={{10}^{4}}vt=8\times {{10}^{3}}t \\
T\left( t \right)&={{T}_{0}}-\frac{{{T}_{0}}-{{T}_{1000}}}{1000}vt=25-1.84\times {{10}^{-2}}t
\end{align}
\end{subequations}
where $v$ is the heave speed of the underwater robot, $T_{0}$ is the water surface temperature, and $T_{1000}$ is the underwater temperature at the depth of 1000 m. Based on the parameters shown in Table \ref{tbl2}, the internal pressure of the joint $p_j$ is solved as:
\begin{equation}\label{eq8}
    {{p}_{j}}\left( t \right)=8000t+2.45\times {{10}^{7}}{{e}^{-1300t}}\left[ \cos \left( 9098t \right)+0.1429\sin \left( 9098t \right) \right]+1.98\times {{10}^{4}}
\end{equation}
\begin{table}[width=.9\linewidth,cols=4,pos=b]
\caption{Parameters of the pressure compensation system.}\label{tbl2}
\begin{tabular*}{\tblwidth}{@{} LLLL@{} }
\toprule
Parameter & Definition & Value & Unit\\
\midrule
$k$ & spring stiffness & \num{2310} & N/m \\
$c$ & coefficient of viscous friction & \num{13000} & $\text{N}\cdot\text{s}$/m \\
$m_c$ & mass of the piston & \num{5} & kg \\
$K$ & bulk modulus of the silicone oil & \num{1034.25e6} & Pa \\
$V_0$ & initial volume of the non-rod chamber and the joint system & \num{3e-3} & $\text{m}^3$ \\
$x_0$ & spring precompression & \num{0.32} & m \\
$S_r$ & effective area of the rolling diaphragm & \num{0.035} & $\text{m}^2$ \\
$\alpha$ & thermal expansion coefficient of the silicone oil & \num{9.5e-4} & 1/C$^{\circ}$ \\
$v$ & heave speed of the underwater robot & \num{0.8} & m/s \\
$T_0$ & water surface temperature & $25$ & C$^{\circ}$ \\
$T_{1000}$ & underwater temperature at the depth of 1000 m & $2$ & C$^{\circ}$ \\
\bottomrule
\end{tabular*}
\end{table}
Fig. \ref{fig_pj} shows the graph of $p_j$. With a heave speed of 0.8 m/s, the underwater robot reaches 1000 m in depth after $t = 1250$ s, where the ambient pressure $p_w$ is 10 MPa. According to Eq (\ref{eq8}), the pressure difference between the inside and the outside $\Delta p$ is:
\begin{equation}\label{eq9}
\Delta p={{p}_{j}}-{{p}_{w}}=\text{0}\text{.01967 MPa}
\end{equation}

Despite the protection against the external liquid incursion, the pressure difference would lead to the leakage of the internal oil, impairing the performance of the pressure compensation and causing pollution. To deal with it, the sealing system of the oil-filled joint should be optimized. Installed between the harmonic drive and the output shaft, rotary lip seals play a pivotal role in the joint system. To boost the sealing performance, we will analyze the principle of the rotary lip seal and extract sealing metrics for mechanical design optimization.

The contact pressure is a key metric in terms of the static seal. Based on an interference fit, the rotary lip seal is assembled between the shaft and the harmonic drive, whose schematics are shown in Fig. \ref{fig_schematic}-(b) and Fig. \ref{fig_schematic}-(c). When the shaft is stationary, the contact pressure between the seal and the shaft causes the deformation of the seal lip made of flexible rubber material. It blocks the gap on the shaft, thereby preventing the leakage of the internal silicone oil. To ensure the effectiveness of the static seal, the contact pressure between the seal lip and the shaft $p_c$ should exceeds the aforementioned pressure difference in Eq (\ref{eq9}):
\begin{equation}\label{eq10}
{{p}_{c}}>\Delta p
\end{equation}

\begin{figure}[!b]
\centering
\includegraphics[width=.5\textwidth]{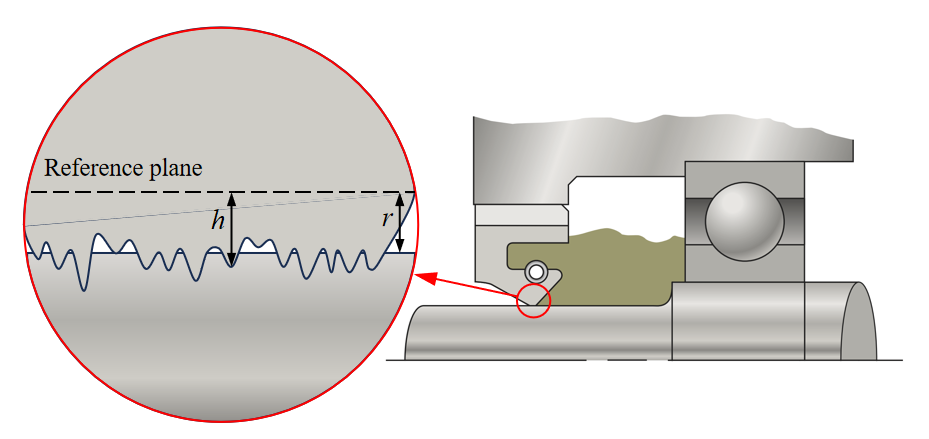}%
\caption{Microscopic model for the static seal.}
\label{fig_seal}
\end{figure}
From a microscopic perspective shown in Fig. \ref{fig_seal}, the contact pressure is calculated based on the Greenwood-Williamson theory \citep{maaboudallah2022review}. We assume that the seal lip surface is an elastic rough plane while the shaft surface is a rigid smooth plane. According to the Greenwood-Williamson theory, the contact pressure between these two planes is related to the asperities distributed along the surface of the seal lip. It is assumed that the height $h$ of an asperity follow a Gaussian distribution:
\begin{equation}
f\left( h \right)=\frac{1}{\sigma }{{e}^{-{{h}^{2}}/2{{\sigma }^{2}}}}
\end{equation}
where $\sigma$ is the standard deviation of the asperity heights. The probability of contact between an asperity of the seal lip and the shaft profile is given by:
\begin{equation}
P\left( h>r \right)=\int\limits_{r}^{\infty }{f\left( h \right)dh}=\frac{1}{\sigma }\int\limits_{r}^{\infty }{{{e}^{-{{h}^{2}}/2{{\sigma }^{2}}}}dh}
\end{equation}
where $r$ is the distance between the reference plane of the asperity height and the shaft profile. Treating the asperity as a spherical tip of a constant curvature, Hertz theory gives the resulting contact force $F$ and contact area $S$ as:
\begin{subequations}\label{eq13}
\begin{align}
    F&=\frac{\sqrt{2}}{3}\frac{{{\rho }^{1/2}}E}{\sigma }\int\limits_{r}^{\infty }{{{\delta }^{3/2}}{{e}^{-{{h}^{2}}/2{{\sigma }^{2}}}}dh}\\
    S&=\frac{\pi \rho \delta P\left( h>r \right)}{2}=\frac{\pi \rho }{2\sigma }\int\limits_{r}^{\infty }{\delta {{e}^{-{{h}^{2}}/2{{\sigma }^{2}}}}dh}
\end{align}
\end{subequations}
where $\rho$ is the radius of curvature, $E$ is the composite Young modulus, and $\delta$ is the interference:
\begin{subequations}
\begin{align}
    \frac{1}{E} &= \frac{1-\nu _{1}^{2}}{{{E}_{1}}}+\frac{1-\nu _{2}^{2}}{{{E}_{2}}}\\
    \delta &= 2(h-r)
\end{align}
\end{subequations}
where $E_1$ and $E_2$ the Young moduli of the seal lip and the shaft, respectively. $\nu_1$ and $\nu_2$ are the corresponding Poisson coefficients. Based on Eq (\ref{eq13}), the contact pressure $p_c$ is given by:
\begin{equation}\label{eq15}
{{p}_{c}}=\frac{F}{S}=\frac{2\sqrt{2}}{3}\frac{{{\rho }^{-1/2}}E}{\pi }\int\limits_{d}^{\infty }{{{\delta }^{1/2}}dh}
\end{equation}
For an asperity with a deterministic height $h$, the expressions of $F$, $S$ and $p_c$ are converted to:
\begin{subequations}
\begin{align}
    F&=\frac{\sqrt{2}{{\rho }^{1/2}}E{\delta }^{3/2}}{3}\\
    S&=\frac{\pi \rho \delta}{2}\\
    {{p}_{c}}&=\frac{2\sqrt{2}}{3}\frac{{{\rho }^{-1/2}}E}{\pi }{{\delta }^{1/2}}
\end{align}
\end{subequations}

In terms of the dynamic seal, reverse pumping mechanism is a classical theory to explain the principle\citep{HUANG2022107917}. When the shaft rotates, there is also axial reciprocating motion between the seal and the shaft due to the manufacturing and assembly tolerance. Under the influence of the contact pressure, the seal lip would experience asymmetric axial friction because of the asymmetric trapezoidal lip structure. Subsequently, the asperities on the seal lip will undergo tangential deformation and generate an oil film, where the hydraulic oil will flow from the air side towards the oil side. Therefore, the rotating shaft lip seal can achieve dynamic sealing.

In the dynamic seal, the pumping rate, which stands for the volumetric flow rate in the oil film, is used to evaluate the sealing performance. Based on the one-dimensional Reynolds equation, the volumetric flow rate $q$ is expressed as:
\begin{equation}\label{eq17}
q=2\pi R\int\limits_{0}^{{{h}_{o}}}{vdy}=2\pi R\int\limits_{0}^{{{h}_{o}}}{\left[ \frac{u\left( {{h}_{o}}-y \right)}{{{h}_{o}}}-\frac{y\left( {{h}_{o}}-y \right)}{2\mu }\frac{dp}{dx} \right]dy}=\pi R\left( -\frac{h_{o}^{3}}{6\mu }\frac{dp}{dx}+u{{h}_{o}} \right)  
\end{equation}
where $R$ is the radius of the shaft, $h_o$ is the height of the oil film, $v$ is the axial velocity of the film fluid, whose radial position is $y$, $u$ is the reciprocating velocity, and $p$ is the pressure of the fluid in the axial direction $x$. Assume the magnitude and direction variation of the reciprocating velocity $u$ is uniform in each rotation cycle, Eq (\ref{eq17}) could be simplified by \citep{kang2015finite}:
\begin{equation}\label{eq18}
q=-\frac{\pi Rh_{o}^{3}}{6\mu }\frac{dp}{dx}
\end{equation}

According to the trapezoidal lip, the axial pressure in the film is set to follow a distribution as:
\begin{equation}\label{eq19}
p(x)=\left\{\begin{matrix} 
&{{p}_{\max }}\frac{x}{{{x}_{\max }}};\text{ }0\le x\le {{x}_{\max }}  \\ 
\\
&{{p}_{\max }}\frac{w-x}{w-{{x}_{\max }}};\text{ }{{x}_{\max }}<x\le w  \\
\end{matrix}\right. 
\end{equation}
where $p_{max}$ is the maximum pressure, and $w$ is the width of the lip. It is assumed that ${p}_{\max }$ occurs at the intersection point of the extended line from the trapezoidal lip's profile:
\begin{equation}
{{x}_{\max }}=\frac{\tan \beta }{\tan \alpha +\tan \beta }w
\end{equation}
where $\alpha$ is the scraper angle and $\beta$ is the barrel angle. Based on the integral of $p(x)$, the load per length experienced by the oil film could be expressed as:
\begin{equation}\label{eq21}
{{F}_{x}}=\int\limits_{0}^{w}{p\left( x \right)dx}=\frac{w{{p}_{\max }}}{2}
\end{equation}
Plug Eq (\ref{eq19}) and Eq (\ref{eq21}) into Eq (\ref{eq18}), the pumping rate could be expressed as:
\begin{equation}
q=-\frac{\pi Rh_{o}^{3}{{F}_{x}}}{3\mu {{w}^{2}}}\left( \frac{{{\tan }^{2}}\alpha -{{\tan }^{2}}\beta }{\tan \alpha \tan \beta } \right)
\end{equation}
The negative sign of the pumping rate denotes the flow direction from the air side to the oil side. When $\lvert q \rvert$ is small, the reverse pumping is subtle, leading to failure of the dynamic seal. On the other hand, when $\lvert q \rvert$ is too large, the oil film between the contact surfaces may break, leading to wear of the seal lip. To evaluate the wear, the frictional heat is defined as:
\begin{equation}
Q=\frac{{{\pi }^{2}}fn{{R}}{{F}{}}}{30}
\end{equation}

The mechanical structure of the joint exerts an influence on the sealing. As a key parameter, the interference between the shaft and the seal lip can be adjusted to improve the sealing performance. To evaluate the effect, the contact pressure $p_c$, the pumping rate $q$, and the frictional heat $Q$ are selected as the indicators. Initially, they are converted to forms with a variable of the interference $\delta$:
\begin{subequations}
\begin{align}
{{p}_{c}}\left( \delta  \right)&=\frac{2\sqrt{2}}{3}\frac{{{\rho }^{-1/2}}E}{\pi }{{\delta }^{1/2}}\\
q&=-\frac{\pi (2R_0+\delta)h_{o}^{3}{{F}_{x}}}{6\mu {{w}^{2}}}\left( \frac{{{\tan }^{2}}\alpha -{{\tan }^{2}}\beta }{\tan \alpha \tan \beta } \right)
\\ 
Q\left( \delta  \right)&=\frac{\sqrt{2}{{\pi }^{2}}{{\rho }^{1/2}}fnE\left( 2{{R}_{0}}+\delta  \right){{\delta }^{3/2}}}{180}\\
\end{align}
\end{subequations}

When the shaft is stationary, Eq (\ref{eq10}) should be met to achieve the static seal. Based on this requirement, we should try to reduce $p_c$ to limit the resulting friction. When the shaft rotates, $\lvert q \rvert$ should be large enough to ensure significant reverse pumping for the dynamic seal. Meanwhile, the frictional heat $Q$ should be constrained to prevent excessive wear.

\begin{figure}[!t]
\centering
\subfloat[]{\includegraphics[width=.35\textwidth]{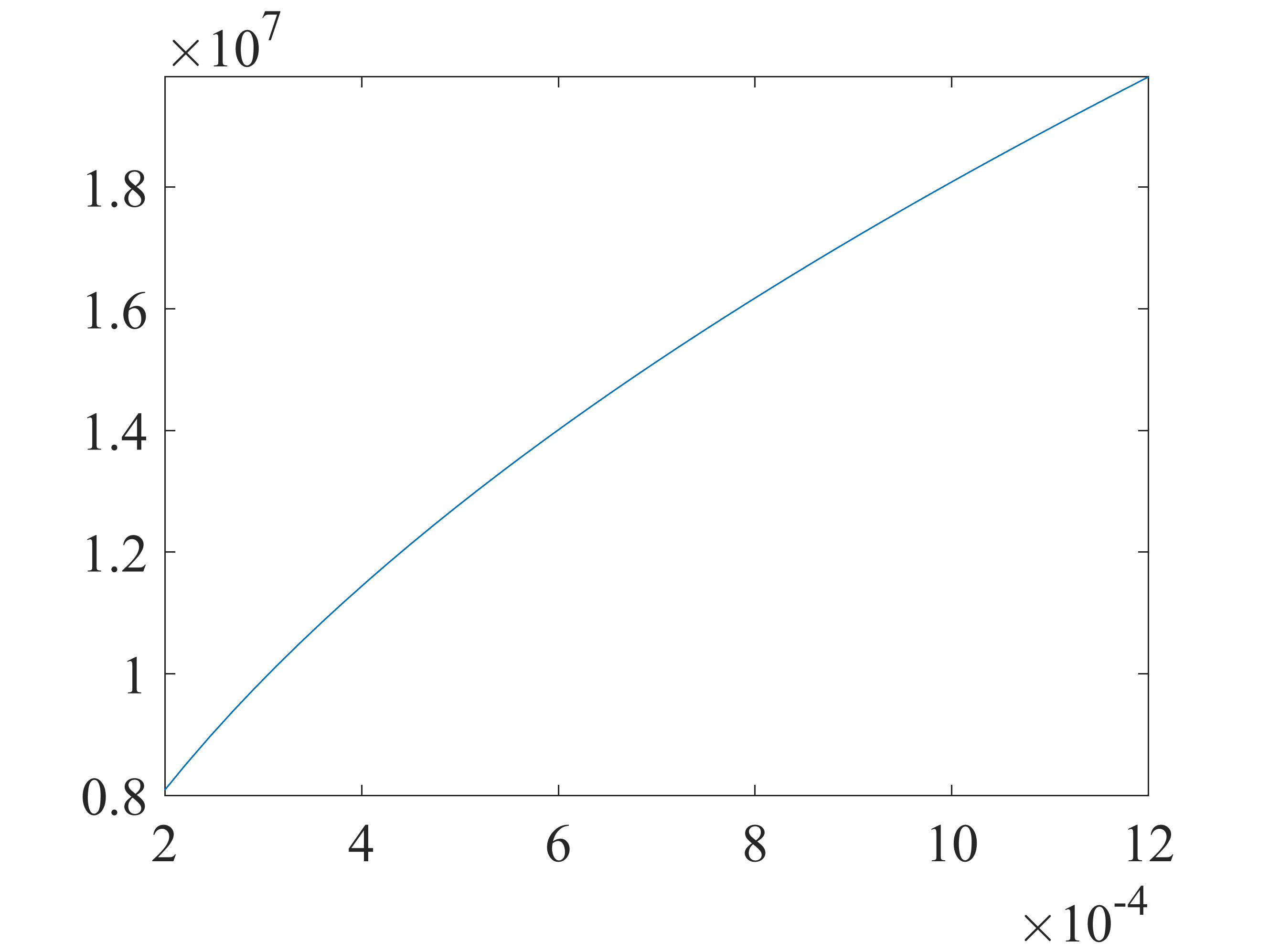}
}
\hfil
\subfloat[]{\includegraphics[width=.35\textwidth]{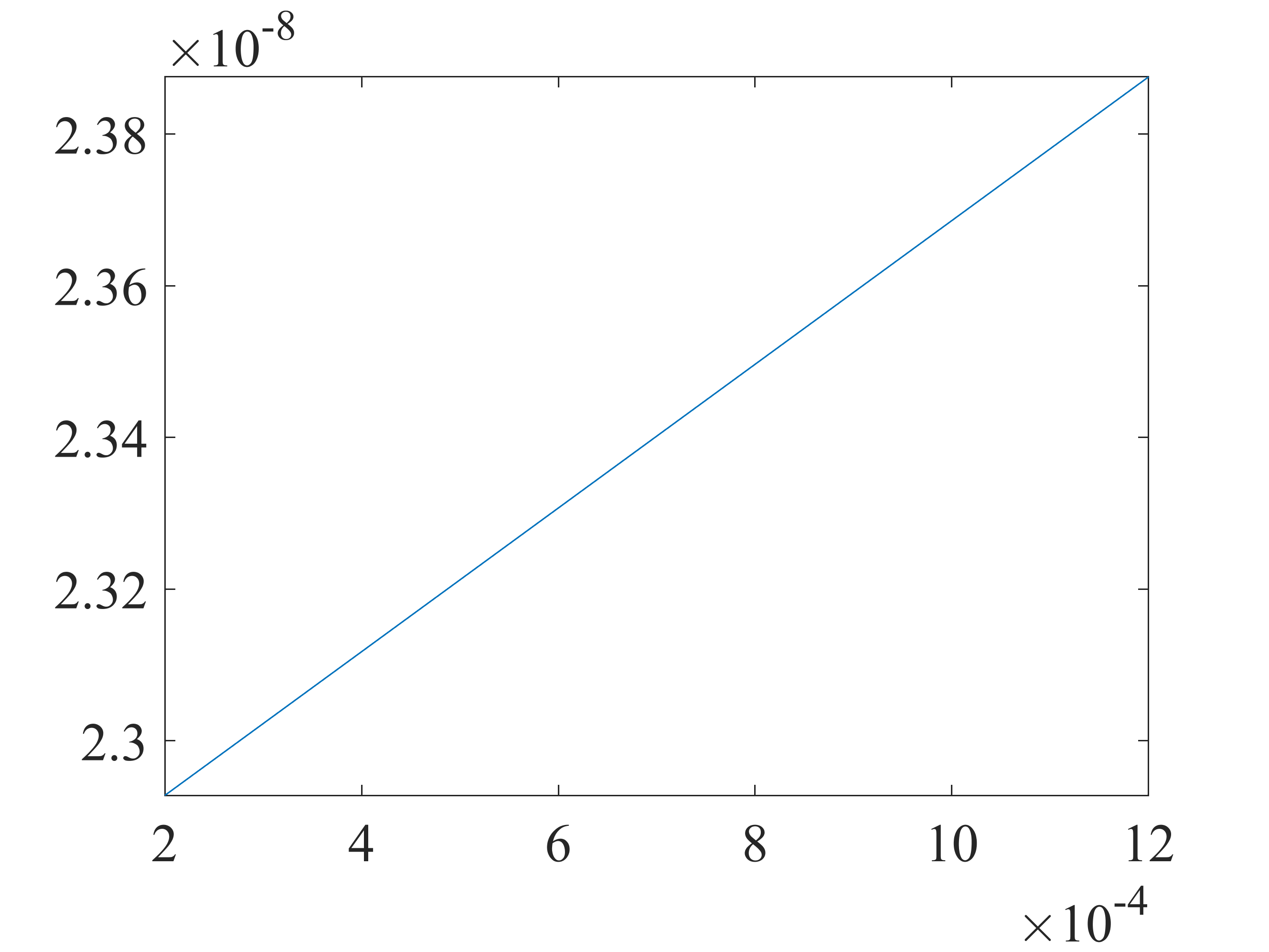}
}
\hfil
\subfloat[]{\includegraphics[width=.35\textwidth]{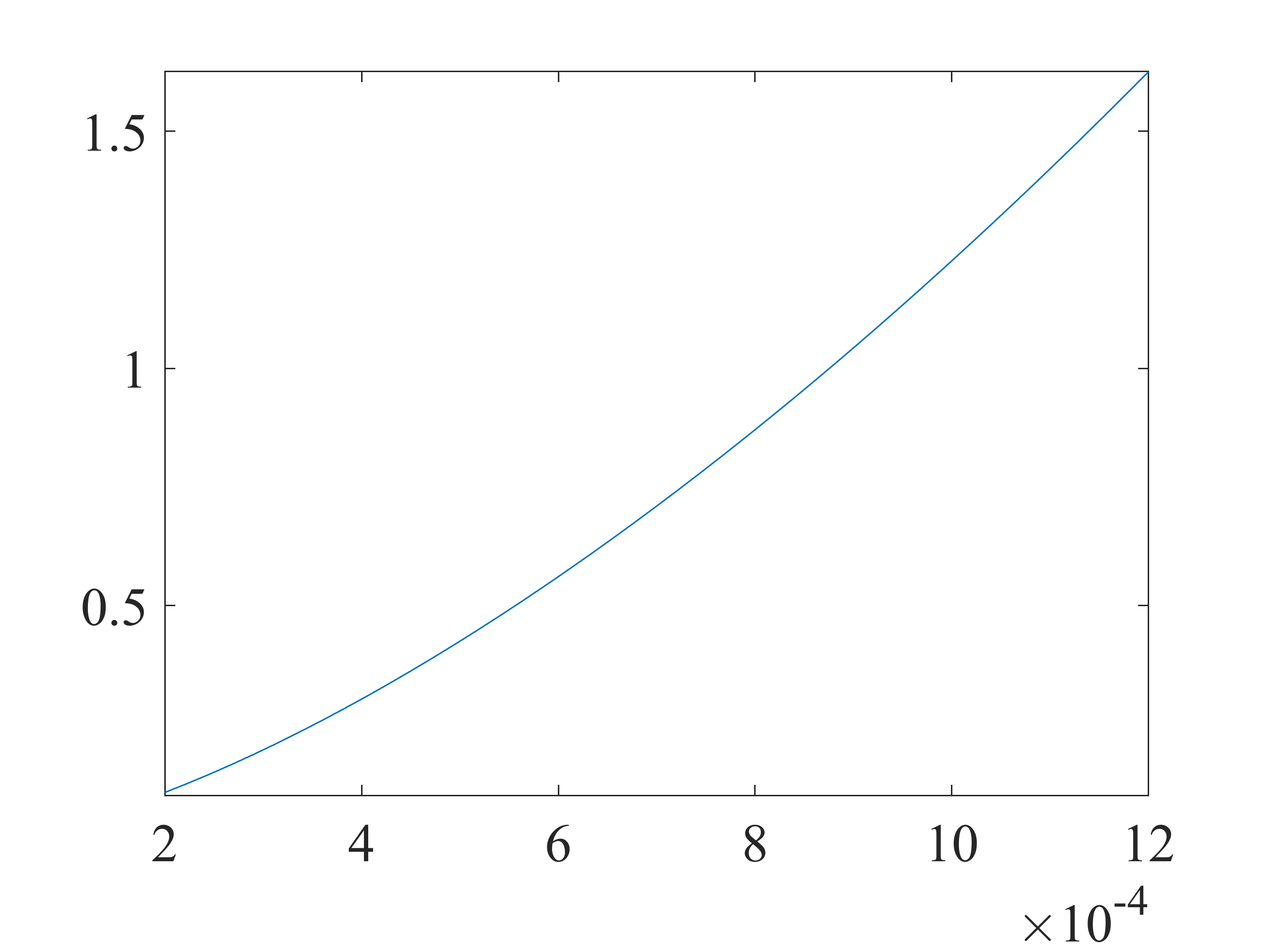}
}
\hfil
\subfloat[]{\includegraphics[width=.35\textwidth]{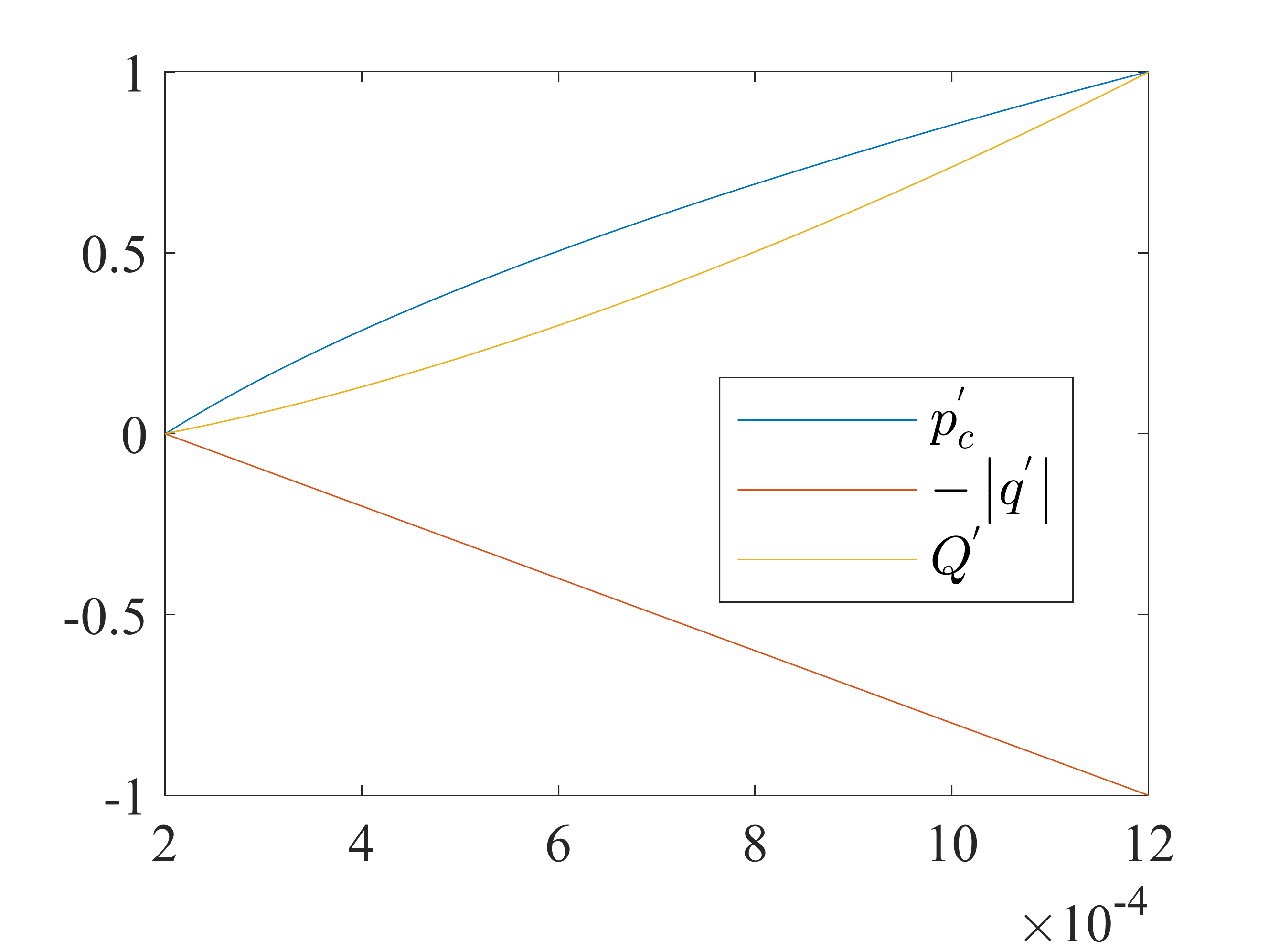}
}
\caption{Graphs of the indicators within the range of the feasible region. (a) Contact pressure. (b) Pumping rate. (c) Frictional heat. (d) Normalization of the indicators.}
\label{fig_fun}
\end{figure}

To comprehensively consider these three indicators, we deploy a multi-objective optimization to search for an optimal interference that meets the requirements. Since the magnitude and unit vary from indicator to indicator, we first normalize them. The feasible region of the interference is empirically defined as $\delta = [0.2,1.2]$ mm. As shown in Fig. \ref{fig_fun}, all of these three indicators are monotonic within this range. Therefore, we can determine the maximum and minimum values of these indicators within this range and then normalize them as follows:
\begin{subequations}
\begin{align}  
{{{p}'}_{c}}\left( \delta  \right)&=\frac{{{p}_{c}}\left( \delta  \right)-\min {{p}_{c}}}{\max {{p}_{c}}-\min {{p}_{c}}}=\frac{{{p}_{c}}\left( \delta  \right)-0.81\times {{10}^{7}}}{1.17\times {{10}^{7}}}\\
\left|{q}'\left( \delta  \right)\right|&=\frac{\left|q\left( \delta  \right)\right|-\min \left|q\right|}{\max \left|q\right|-\min \left|q\right|}=\frac{\left|q\left( \delta  \right)\right|-2.29\times {{10}^{-8}}}{0.095\times {{10}^{-8}}}\\
{Q}'\left( \delta  \right)&=\frac{Q\left( \delta  \right)-\min Q}{\max Q-\min Q}=\frac{Q\left( \delta  \right)-0.11}{1.52}
\end{align}
\end{subequations}

Based on the normalization, the multi-objective optimization is defined as follows:
\begin{equation}
\begin{aligned}
\underset{\delta }{\mathop{\min }}\,f\left( \delta  \right)&={{\omega }_{1}}{{{{p}'}}_{c}}\left( \delta  \right)-{{\omega }_{2}}\left| {q}'\left( \delta  \right) \right|+{{\omega }_{3}}{Q}'\left( \delta  \right) \\ 
\text{s}\text{.t}\text{. }\delta &\in \left[ 0.2\times {{10}^{-3}},1.2\times {{10}^{-3}} \right]\text{m}\\ 
\text{      }{{p}_{c}}&>\Delta p \\ 
\text{      }q\left( \delta  \right)&>4\times {{10}^{-10} \text{m}^3}\text{/s} \\ 
\end{aligned}
\end{equation}

\begin{figure}[!h]
\centering
\includegraphics[width=.35\textwidth]{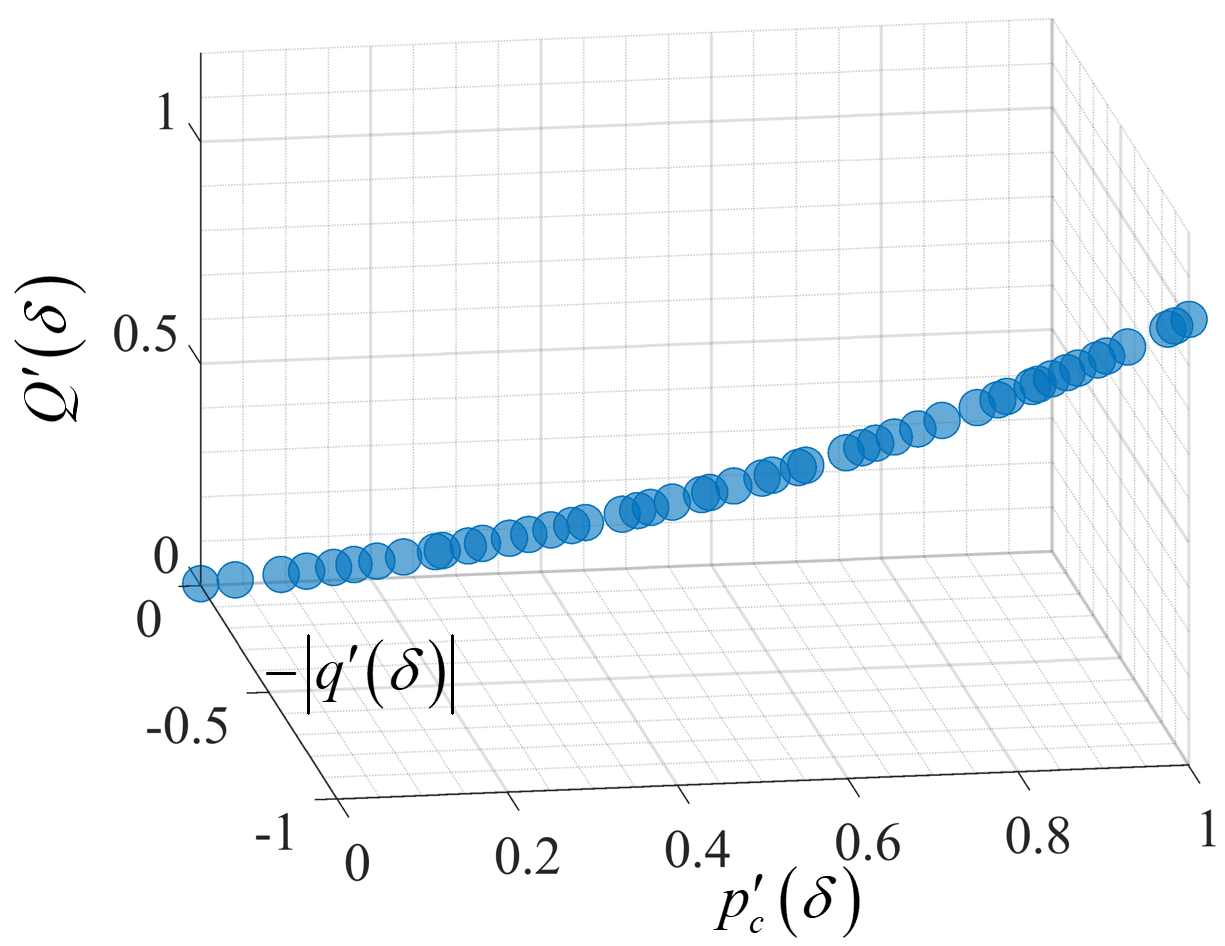}%
\caption{Resulting Pareto front.}
\label{fig_pareto}
\end{figure}

For this nonlinear, single-variable, constrained multi-objective optimization problem, the Non-Dominated Sorting Genetic Algorithm II (NSGA-II) is applied to solve it, yielding a Pareto optimal set as shown in Fig. \ref{fig_pareto}. From this set, $\delta^* = 0.627$ mm is selected to be the optimized interference. Substituting this value into the constraint functions, it confirms that constraints are met. For different joints, we could just adjust the mechanical parameters and follow the same procedure to implement the optimal design.

\begin{table}[width=.9\linewidth,cols=4,pos=t]
\caption{Mechanical parameters of the rotary lip seal.}\label{tbl3}
\begin{tabular*}{\tblwidth}{@{} LLLL@{} }
\toprule
Parameter & Definition & Value & Unit\\
\midrule
$E_1$ & Young modulus of the seal lip & \num{5.54} & MPa \\
$E_2$ & Young modulus of the shaft & \num{2.0e5} & MPa \\
$\nu_1$ & Poisson coefficient of the seal lip & \num{0.4995} & - \\
$\nu_2$ & Poisson coefficient of the shaft & \num{0.29} & - \\
$\rho$ & radius of curvature of the asperity & \num{15} & $\mu\text{m}$ \\
$\alpha$ & scraper angle & $\pi/4$ & rad \\
$\beta$ & barrel angle & $\pi/9$ & rad \\
$R_0$ & nominal radius of the shaft & \num{12} & mm \\
$f$ & coefficient of friction & \num{0.4} & - \\
$w$ & contact width of the seal lip & $0.2$ & mm \\
$n$ & angular velocity of the shaft & $1750$ & rpm \\
$u$ & reciprocating velocity of the shaft & \num{6e-3} & $\text{m/s}$ \\
$F_x$ & load per length of the seal film & $300$ & N/m \\
$h_{o}$ & thickness of the oil film & $2$ & $\mu\text{m}$ \\
\bottomrule
\end{tabular*}
\end{table}

\section{Robust position control of oil-filled joint}\label{sec4}
As mentioned before, to withstand the substantial underwater pressure, the joint system is filled with silicone oil, maintaining an internal pressure nearly equal to that of the ambient one. However, the silicone oil complicates the original dynamic model of the joint system and introduces unknown disturbances. Additionally, the viscous oil deteriorates the wear of the motor, causing parametric uncertainty. To address the dynamic perturbations and disturbance signals, we first establishes the dynamic model of the oil-filled underwater electric joint. Based on this model, a robust position controller is designed using the structured singular value synthesis ($\mu$ synthesis) method.

The nominal dynamic model of the oil-filled joint starts with the electromechanical equation of a PMSM in a rotor
rotating reference frame $d$-$q$:
\begin{subequations}\label{27}
\begin{align}
  J{{\dot{\omega }}_{m}}&={{T}_{e}}-{{T}_{l}}-{{k}_{b}}{{\omega }_{m}}\\
  {{T}_{e}}&={{k}_{t}}{{i}_{q}}
\end{align}
\end{subequations}
where $J$ is the moment of inertia, $\omega_m$ is the rotor velocity, $T_e$ is the electromagnetic torque, $T_l$ is the load torque, $k_b$ is the viscous damping, $k_t$ is the torque constant, and $i_q$ is the $q$-axis stator current in the $d$-$q$ coordinate, standing for the decoupled torque current. Based on the substitution of joint output position $\theta_j$ for $\omega_m$, Eq (\ref{27}) is converted to:
\begin{equation}
    {{\ddot{\theta }}_{j}}=-\frac{{{k}_{b}}}{J}{{\dot{\theta }}_{j}}+\frac{{{k}_{t}}}{JI}{{i}_{q}}-\frac{1}{JI}{{T}_{l}}
\end{equation}
where $I$ is the reduction ratio of the harmonic drive. Considering the joint filled with viscous oil, the dynamic model is supplemented by:
\begin{equation}\label{eq29}
    {{\ddot{\theta }}_{j}}=-\frac{{{k}_{b}}}{J}{{\dot{\theta }}_{j}}+\frac{{{k}_{t}}}{JI}{{i}_{q}}-\frac{1}{JI}{{T}_{l}}-\frac{1}{JI}\left( {{T}_{o}}+\frac{{{T}_{s}}}{I}+\frac{{{T}_{d}}}{I} \right)
\end{equation}
where $T_o$, $T_s$, and $T_d$ are the oil stirring loss torque, dynamic seal loss torque, and unknown external disturbance torque, respectively.

\begin{figure}[!t]
   \centering
   \includegraphics[width=.4\textwidth]{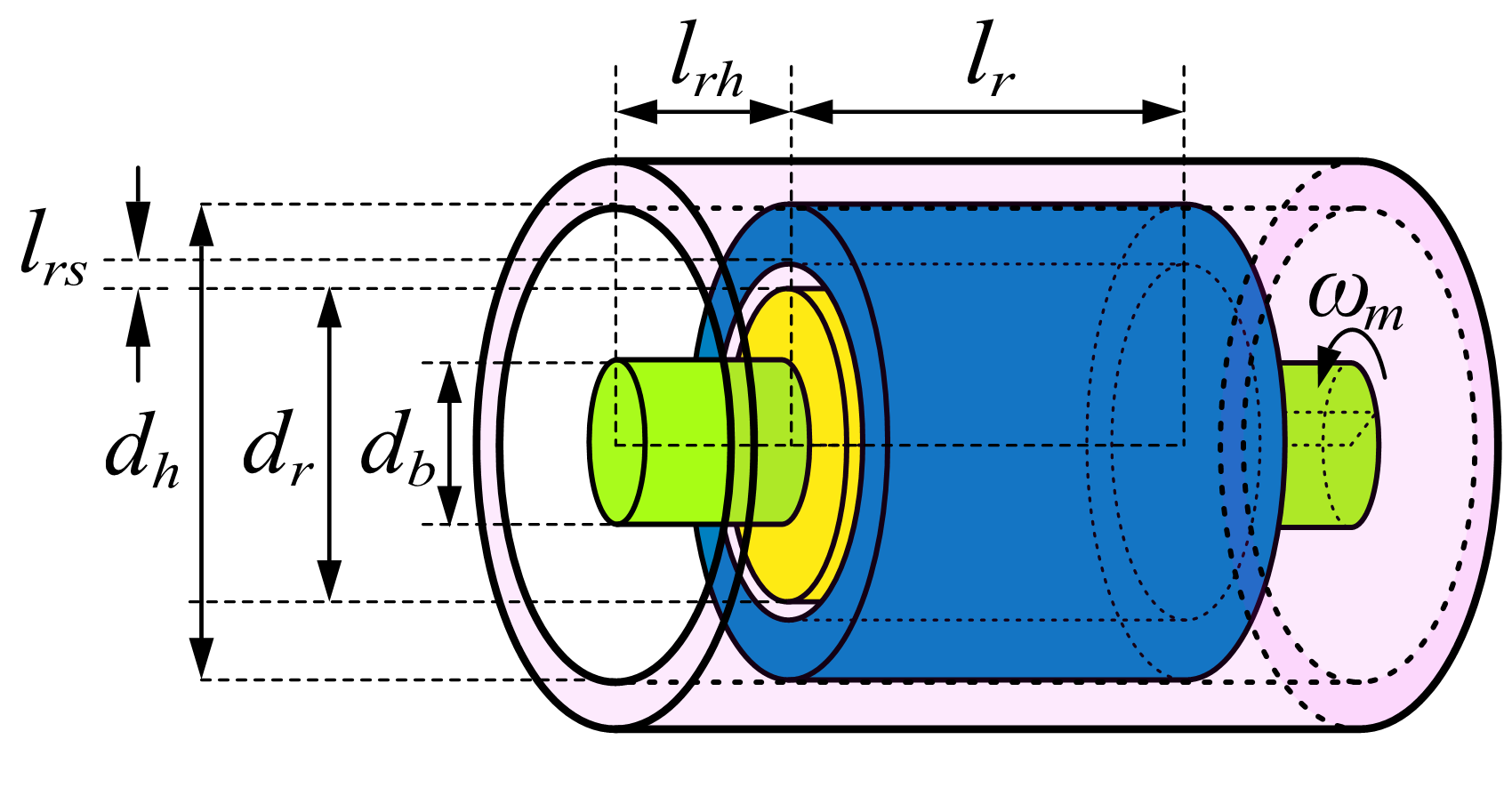}
   \caption{Simplified joint structure.}
   \label{fig_simplified}
\end{figure}

Due to the viscosity of silicone oil, the motor must overcome viscous friction resistance during rotation, leading to corresponding losses. The oil stirring loss torque of the joint comprises the disk loss torque between the rotor and the stator, as well as the flank loss torque between the rotor and the joint housing (or bearing). Initially, the structure of the joint is simplified, and we assume that the oil is Newtonian fluid in the laminar state. Fig. \ref{fig_simplified} illustrates the simplified joint structure.

In terms of the disk loss torque, the velocity gradient of the oil between the outer rotor and inner stator is:
\begin{equation}
    \nabla v=\frac{{{\omega }_{m}}{{d}_{r}}}{2{{l}_{rs}}}
\end{equation}
where $d_r$ is the outer diameter of the rotor, and $l_{rs}$ is the radial clearance between the rotor and the stator. By Newton’s internal friction law, the viscous shear stress is expressed as:
\begin{equation}
    \tau =\mu \nabla v=\frac{\mu {{\omega }_{m}}{{d}_{r}}}{2{{l}_{rs}}}
\end{equation}
The corresponding viscous friction on the outer rotor is:
\begin{equation}
    {{F}_{v}}=\tau \cdot \pi {{d}_{r}}{{l}_{r}}=\frac{\pi \mu {{\omega }_{m}}{{l}_{r}}{{d}_{r}}^{2}}{2{{l}_{rs}}}
\end{equation}
where $l_r$ is the axial length of the rotor. Hence, the disk loss torque is given by:
\begin{equation}
    {{T}_{v}}={{F}_{v}}\frac{{{d}_{r}}}{2}=\frac{\pi \mu {{\omega }_{m}}{{l}_{r}}{{d}_{r}}^{3}}{4{{l}_{rs}}}
\end{equation}

The flank torque emerges between the end faces of a rotor and a steady housing component connected by a shaft with a diameter of $d_b$. For a differential ring element $dr$ at radius $r$ on the rotor end face, the velocity gradient of the oil between the element and the housing component is expressed as:
\begin{equation}
\nabla {{{v}'}_{r}}=\frac{{{\omega }_{m}}r}{{{l}_{rh}}}
\end{equation}
where $l_{rh}$ is the axial distance between the rotor and the housing component. Based on Newton's internal friction law, the differential of the flank torque is given by:
\begin{equation}\label{eq35}
    d{{T}_{ve}}=d{{F}_{ve}}r=\frac{2\pi \mu {{\omega }_{m}}{{r}^{3}}}{{{l}_{rh}}}dr
\end{equation}
The total flank torque is obtained based on integrating Eq (\ref{eq35}) between the axis and the outer rotor:
\begin{equation}
    {{T}_{ve}}=\int\limits_{{{d}_{b}}/2}^{{{d}_{r}}/2}{d{{T}_{ve}}}=\int\limits_{{{d}_{b}}/2}^{{{d}_{r}}/2}{\frac{2\pi \mu {{\omega }_{m}}{{r}^{3}}}{{{l}_{rh}}}dr}=\frac{\pi ({{d}_{r}}^{4}-{{d}_{b}}^{4})\mu {{\omega }_{m}}}{32{{l}_{rh}}}
\end{equation}

\begin{figure}[!t]
   \centering
   \includegraphics[width=.6\textwidth]{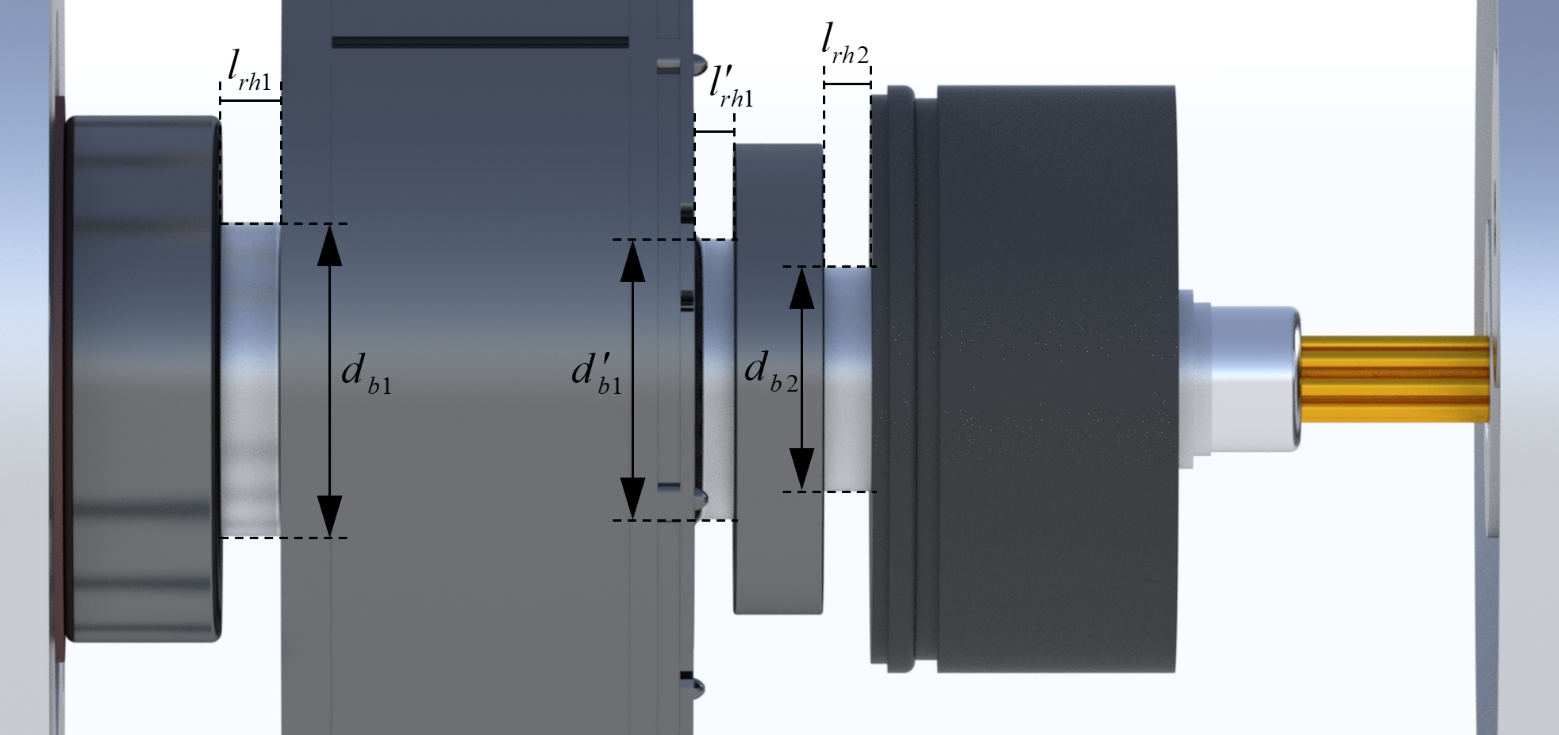}
   \caption{Dimensions of the clearance.}
   \label{fig_clearance}
\end{figure}

As mentioned in Section \ref{sec2}, the joint system contains a motor and a resolver, both of which contain a stator and a rotor. Therefore, the total oil-stirring loss torque should be expressed as the sum of the disk and flank torques from both the motor and the resolver. For the motor, it's equipped with bearings at both ends. They are seen as the steady housing components. For the resolver, one of its end is next to a bearing, while the other is next to a end cap. Since the distance between the rotor and the end cap is relatively great, the relevant flank torque is neglected. To sum up, the total oil-stirring loss torque is expressed by:
\begin{equation}\label{eq37}
\begin{aligned}
  {{T}_{o}}&={{T}_{v1}}+{{T}_{v2}}+{{T}_{ve1}}+{{{{T}'}}_{ve1}}+{{T}_{ve2}} \\ 
 & =\left[ \frac{\pi {{l}_{r1}}{{d}_{r1}}^{3}}{4{{l}_{rs1}}}+\frac{\pi {{l}_{r2}}{{d}_{r2}}^{3}}{4{{l}_{rs2}}}+\frac{\pi ({{d}_{r1}}^{4}-{{d}_{b1}}^{4})}{32{{l}_{rh1}}}+\frac{\pi ({{d}_{r1}}^{4}-{{{{d}'}}_{b1}}^{4})}{32{{{{l}'}}_{rh1}}}+\frac{\pi (d_{r2}^{4}-d_{b2}^{4})}{32{{l}_{rh2}}} \right]\mu {{\omega }_{m}} \\ 
 & =\left[ \frac{\pi {{l}_{r1}}{{d}_{r1}}^{3}}{4{{l}_{rs1}}}+\frac{\pi {{l}_{r2}}{{d}_{r2}}^{3}}{4{{l}_{rs2}}}+\frac{\pi ({{d}_{r1}}^{4}-{{d}_{b1}}^{4})}{32{{l}_{rh1}}}+\frac{\pi ({{d}_{r1}}^{4}-{{{{d}'}}_{b1}}^{4})}{32{{{{l}'}}_{rh1}}}+\frac{\pi (d_{r2}^{4}-d_{b2}^{4})}{32{{l}_{rh2}}} \right]I\mu {{\omega }_{j}}  
\end{aligned}
\end{equation}
where $l_{r1}$ and $l_{r2}$ are the rotor's axial lengths of the motor and the resolver, respectively, $l_{rs1}$ and $l_{rs2}$ are the corresponding radial clearances, $d_{r1}$ and $d_{r2}$ are the corresponding rotor's diameters, $d_{b1}$, $d_{b1}^{'}$, $d_{b2}$ are the corresponding diameters of the axis between the rotor and the steady components, and $l_{rh1}$, $l_{rh1}^{'}$, $l_{rh2}$ are the corresponding axial distances. Fig. \ref{fig_clearance} depicts the dimensions of the clearance.

In Eq (\ref{eq29}), apart from the oil stirring loss torque $T_o$, it's difficult to explicitly model the dynamic seal loss torque $T_s$ and the external disturbance torque $T_d$. Besides, the load torque $T_l$ varies from task to task. To address them, we treat the term $(T_l+T_s/I+T_d/I)/JI$ in Eq (\ref{eq29}) as a disturbance signal $d$:
\begin{equation}
    d=\frac{1}{JI}(T_l+\frac{T_s}{I}+\frac{T_d}{I})
\end{equation}
We will try to attenuate it using a robust position controller.

Based on Eq (\ref{eq29}), Eq (\ref{eq37}), and Eq (\ref{eq38}), the state-space representation of the nominal oil-filled joint dynamic model is written as:
\begin{equation}\label{eq38}
\begin{aligned}
\dot{x}&=\left[ \begin{matrix}
   0 & 1  \\
   0 & \frac{-{{k}_{b}}I-{{K}_{d}}}{JI}  \\
\end{matrix} \right]x+\left[ \begin{matrix}
   0  \\
   \frac{{{k}_{t}}}{JI}  \\
\end{matrix} \right]u+\left[ \begin{matrix}
   0  \\
   -d  \\
\end{matrix} \right] \\ 
 y&=\left[ \begin{matrix}
   1 & 0  \\
\end{matrix} \right]x \\ 
\end{aligned}
\end{equation}
where $x=[x_1\,x_2]^\text{T}=[\theta_j\,\dot{\theta_j}]^\text{T}$ is the state vector, $u = i_q$ is the control vector, and $y =x_1= \theta_j$ is the output vector. Based on Eq (\ref{eq37}), the coefficient of oil-stirring loss torque $K_d$ is 
\begin{equation}
{{K}_{d}}=\left[ \frac{\pi {{l}_{r1}}{{d}_{r1}}^{3}}{4{{l}_{rs1}}}+\frac{\pi {{l}_{r2}}{{d}_{r2}}^{3}}{4{{l}_{rs2}}}+\frac{\pi ({{d}_{r1}}^{4}-{{d}_{b1}}^{4})}{32{{l}_{rh1}}}+\frac{\pi ({{d}_{r1}}^{4}-{{{{d}'}}_{b1}}^{4})}{32{{{{l}'}}_{rh1}}}+\frac{\pi (d_{r2}^{4}-d_{b2}^{4})}{32{{l}_{rh2}}} \right]I\mu
\end{equation}

\begin{figure}[!h]
   \centering
   \includegraphics[width=.6\textwidth]{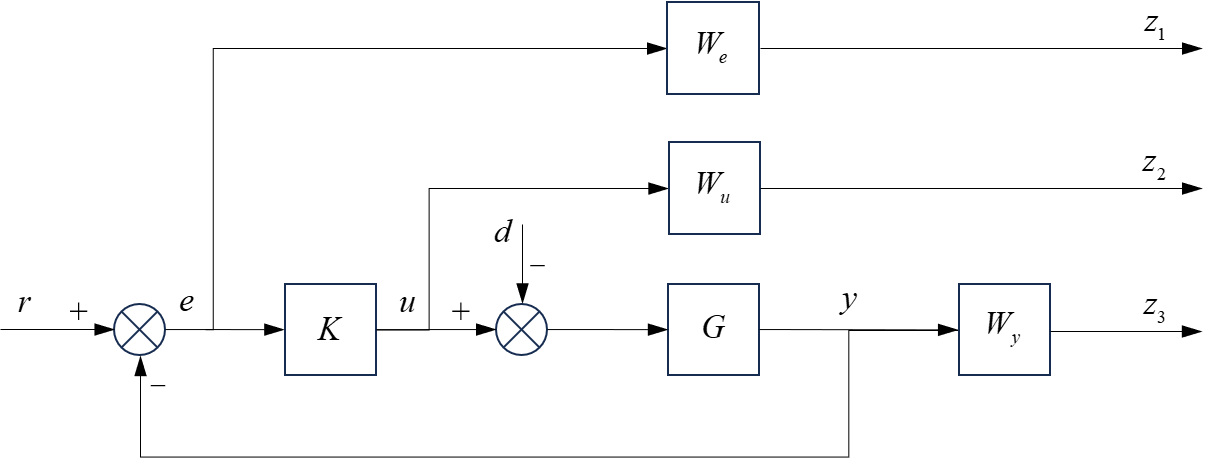}
   \caption{Closed-loop design structure.}
   \label{fig_wf}
\end{figure}

To construct a closed-loop control framework, we establish a design configuration as shown in Fig. \ref{fig_wf}. For the specification of the control performance, weighting functions $W_e$, $W_u$, and $W_y$ are introduced to guide the generation of the controller $K$. They are used to limit the steady-state error, the energy of the control signal, and the overshooting, respectively. Besides, we treat the disturbance $d$ in the plant model $G$ as another input signal for further processing. As the controller output signal $u$ (input signal of $G$) stands for $i_q$, which is proportional to torque, this adjustment is reasonable.

\begin{figure}[!h]
   \centering
   \includegraphics[width=.7\textwidth]{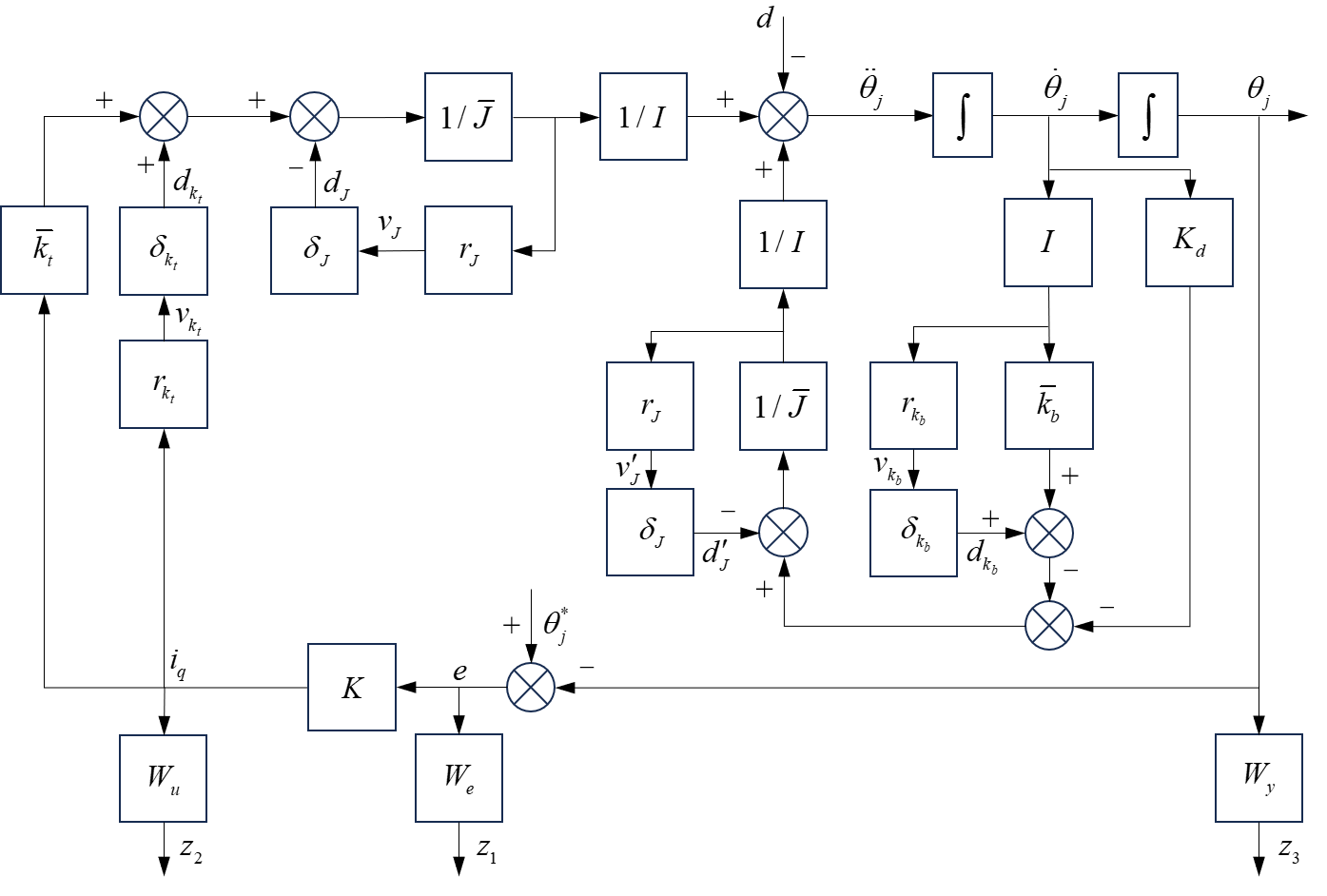}
   \caption{Modified design structure.}
   \label{fig_mds}
\end{figure}

Based on the nominal plant model $G$, we try to address the deteriorated wear-and-tear effect caused by the viscous oil. Such an effect would lead to dynamic perturbations in the system. To consider it, several key parameters in $G$ are written as a combination of the nominal value and the variation:
\begin{equation}\label{eq41}
\begin{aligned}
J&=\bar{J}+{{r}_{J}}{{\delta }_{J}} \\ 
{{k}_{t}}&={{{\bar{k}}}_{t}}+{{r}_{{{k}_{t}}}}{{\delta }_{{{k}_{t}}}} \\ 
{{k}_{b}}&={{{\bar{k}}}_{b}}+{{r}_{{{k}_{b}}}}{{\delta }_{{{k}_{b}}}} \\ 
\end{aligned}
\end{equation}
where $\bar{J}$, $\bar{k}_t$, $\bar{k}_b$ are the nominal values of the moment of inertia, the torque constant and the viscous damping, respectively, $r_J$, $r_{k_t}$, $r_{k_b}$ are the corresponding relative uncertainties ($0 \leq r_a \leq \bar{a}$), and $\delta_J$, $\delta_{k_t}$, $\delta_{k_b}$ are the corresponding normalized uncertainties ($-1 \leq \delta_J, \delta_{k_t}, \delta_{k_b}\leq 1$). With the substitution of the parametric uncertainty in Eq (\ref{eq41}), the design configuration is modified to the form as shown in Fig. \ref{fig_mds}, where ${v}_{\Delta }=[{v}_{J}\,{{{v}'}}_{J}\,{v}_{{{k}_{t}}}\,{v}_{{{k}_{b}}}]^{\text{T}}$ is the input vector of the normalized uncertainties, and ${d}_{\Delta }=[{d}_{J}\,{{{d}'}}_{J}\,{d}_{{{k}_{t}}}\,{d}_{{{k}_{b}}}]^{\text{T}}$
is the corresponding output vector. 

Collecting the normalized uncertainties in a block diagonal matrix, the structured uncertainty is represented as:
\begin{equation}
    \begin{aligned}
\Delta &=\left[ \begin{matrix}
   {{\delta }_{J}} & {} & {} & {}  \\
   {} & {{\delta }_{J}} & {} & {}  \\
   {} & {} & {{\delta }_{{{k}_{t}}}} & {}  \\
   {} & {} & {} & {{\delta }_{{{k}_{b}}}}  \\
\end{matrix} \right] \\ 
{{d}_{\Delta }}&=\Delta {{v}_{\Delta }}  
\end{aligned}
\end{equation}

\begin{figure}[!h]
   \centering
   \includegraphics[width=.25\textwidth]{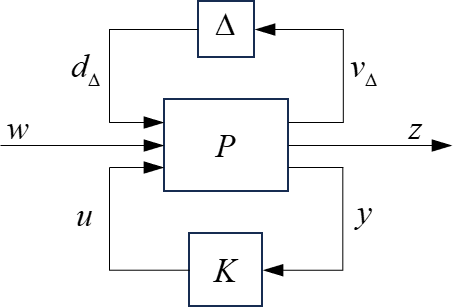}
   \caption{Standard control configuration.}
   \label{fig_scc}
\end{figure}

With the separation of the structured uncertainty $\Delta$ and the controller $K$ from the design configuration, a standard control configuration is established as shown in Fig. \ref{fig_scc}, where $P$ is the generalized plant containing the relative uncertainties. 

The relationship between the input and the output of $P$ is written as:
\begin{equation}
    \left[ \begin{matrix}
   {{v}_{\Delta }}  \\
   z  \\
   y  \\
\end{matrix} \right]=P\left[ \begin{matrix}
   {{d}_{\Delta }}  \\
   w  \\
   u  \\
\end{matrix} \right]
\end{equation}

In this case, the weighting functions are selected as:
\begin{subequations}
\begin{align}
    {{W}_{e}}&=\frac{0.1s+9.95\times {{10}^{-2}}}{\text{   }s+9.95\times {{10}^{-4}}}\\    
    {{W}_{u}}&=\frac{\text{1}{{\text{0}}^{4}}s+\text{7}\text{.538}\times \text{1}{{\text{0}}^{4}}}{s+\text{7}\text{.538}\times \text{1}{{\text{0}}^{5}}}\\
    {{W}_{y}}&=\frac{10s+5}{s+50}
\end{align}
\end{subequations}

The parametric uncertainties are specified by:
\begin{subequations}
\begin{align}
  J&=\bar{J}+{{r}_{J}}{{\delta }_{J}}=\bar{J}+0.2\bar{J}{{\delta }_{J}}=9.6\times {{10}^{-6}}+1.92\times {{10}^{-6}}{{\delta }_{J}} \\ 
 {{k}_{t}}&={{{\bar{k}}}_{t}}+{{r}_{{{k}_{t}}}}{{\delta }_{{{k}_{t}}}}={{{\bar{k}}}_{t}}+0.2{{{\bar{k}}}_{t}}{{\delta }_{{{k}_{t}}}}=0.132+2.64\times {{10}^{-2}}{{\delta }_{{{k}_{t}}}} \\ 
 {{k}_{b}}&={{{\bar{k}}}_{b}}+{{r}_{{{k}_{b}}}}{{\delta }_{{{k}_{b}}}}={{{\bar{k}}}_{b}}+0.2{{{\bar{k}}}_{b}}{{\delta }_{{{k}_{b}}}}=2\times {{10}^{-5}}+4\times {{10}^{-6}}{{\delta }_{{{k}_{b}}}}  
\end{align}
\end{subequations}

Following the standard control configuration, a robust controller could be designed by the $\mu$-synthesis using the DGK-iterations\citep{zhou1998essentials}:
\begin{equation}
    {{K}_{r}}\left( s \right)=\frac{173.7{{s}^{2}}+362.6s+95.88}{{{s}^{3}}+61.22{{s}^{2}}+2430s+2.414}
\end{equation}

\begin{figure}[!t]
\centering
\subfloat[]{\includegraphics[width=.32\textwidth]{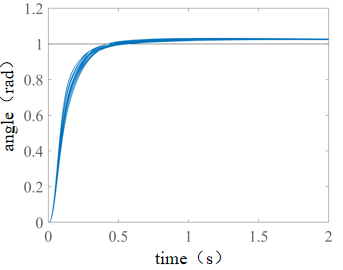}}\label{fig_sr_1}
\hfil
\subfloat[]{\includegraphics[width=.32\textwidth]{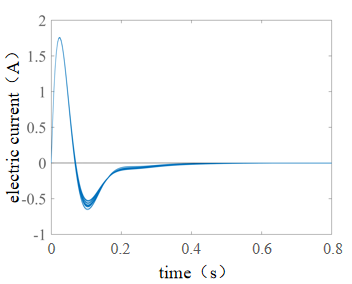}}\label{fig_sr_2}
\hfil
\subfloat[]{\includegraphics[width=.32\textwidth]{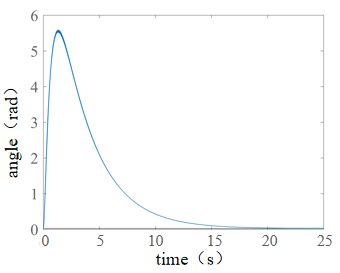}}\label{fig_sr_3}
\hfil
\caption{Step responses of the sensitivity functions. (a) $T_{yr}$. (b) $T_{ur}$. (c) $T_{ed}$.}
\label{fig_step_response}
\end{figure}

To evaluate the performance of $K_r(s)$, 3 sensitivity functions $T_{yr}$, $T_{ur}$, and $T_{ed}$ are analyzed, referring to the ability of signal tracking, energy efficiency, and disturbance attenuation, respectively. We write the closed-loop transfer function $T_{yr}$ from the reference input $r$ to the plant output $y$ as:
\begin{equation}
    {{T}_{yr}}=\frac{GK}{1+GK}
\end{equation}
The unit step response of $T_{yr}$ is shown in Fig. \ref{fig_sr_1} (a). To consider parametric uncertainties, 20 plant models are randomly generated within the variation range of the parameters. In terms of the transient performance, the closed-loop system with $K_r$ achieves a rise time around 0.5 s without overshooting. In the steady-state, the tracking error is around 0.03, which is acceptable for the underwater scenario.

The closed-loop transfer function from the reference input $r$ to the controller output $u$ is represented by $T_{ur}$:
\begin{equation}
    {{T}_{ur}}=\frac{K}{1+GK}
\end{equation}
The relevant unit step response is illustrated in Fig. \ref{fig_sr_2} (b), which indicates that the controller output ($u=i_q$) changes gradually and peaks at 1.75. This value is affordable to the physical plant system.

$T_{ed}$ is the closed-loop transfer function from the disturbance input $d$ to the plant output $y$: 
\begin{equation}
    {{T}_{ed}}=\frac{-G}{1+KG}
\end{equation}
The corresponding step response graph Fig. \ref{fig_sr_3} (c) shows that the closed-loop system could attenuate the disturbance input within a reasonable time. These closed-loop transient responses validate the robust performance of the designed controller.

\section{Experimental validations}\label{sec5}
\begin{figure}[!h]
\centering
\subfloat[]{\includegraphics[width=.32\textwidth]{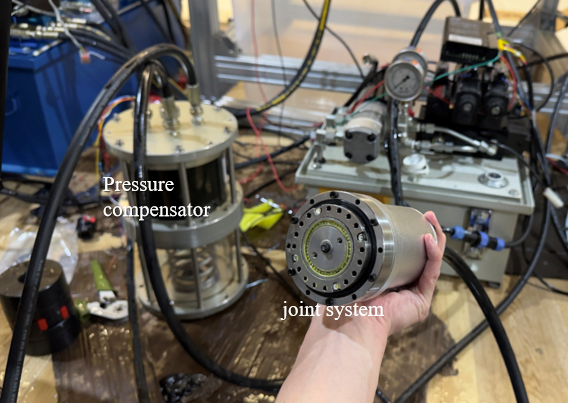}}
\hfil
\subfloat[]{\includegraphics[width=.4\textwidth]{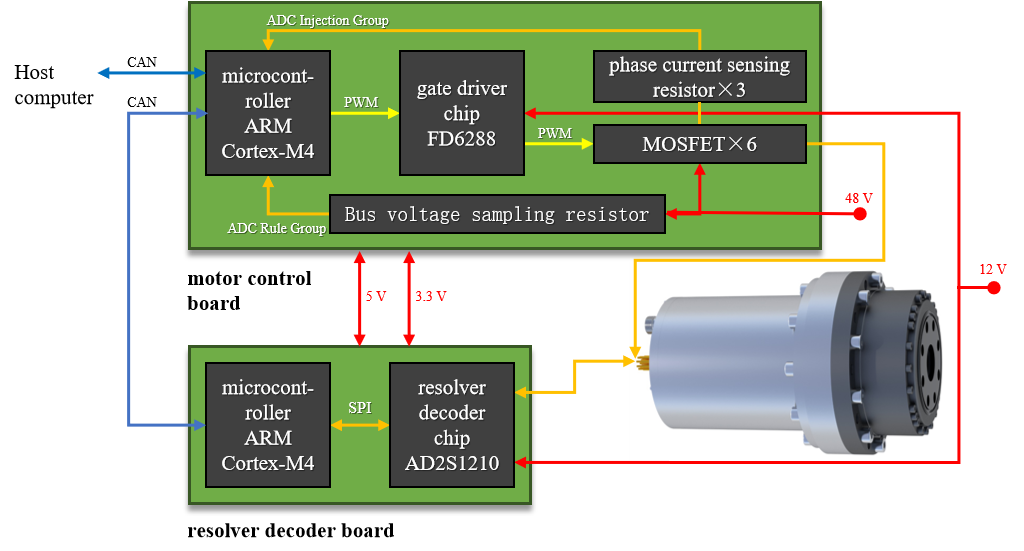}}
\hfil
\caption{Experiment setup. (a) Mechanical module. (b) Drive module.}
\label{physical_modules}
\end{figure}

To verify the performance of the designed robust position controller for the oil-filled electric joint system, relevant experiments are carried out. Based on the mechatronic framework introduced in Section \ref{sec2}, we have developed an electric wrist joint system for an underwater manipulator. The mechanical module mainly comprises a frameless torque motor FMC05707, a harmonic drive with a reduction ratio of 80, and a resolver TS2640N321E64. The drive module is composed of a resolver decoder board and a motor control board. Both boards are based on a ARM Cortex-M4 microcontroller and communicate via CAN (controller area network) bus. The resolver decoder board contains a chip AD2S1210 to obtain the digital motor position from the analog voltage output of the resolver. After transmitting the motor position signal to the motor control board as a feedback, drive and control algorithms are implemented by the motor control board. A sealed connector is utilized to connect the mechanical module with the drive one in the underwater environment. Fig. \ref{physical_modules} shows the mechanical module and drive module, respectively. 

The developed joint system is expected to work at a depth of 1000 m, where the ambient pressure is about 10 MPa. In this underwater environment, the joint will be filled with pressurized silicone oil with a nominal kinematic viscosity of 50 cSt to achieve pressure compensation. However, under such harsh conditions, it's comparatively demanding to conduct control experiments to verify the effect of the filled oil and the performance of the robust controller. Considering that the viscosity rises along with the increase of the depth owing to the change of the ambient pressure and temperature, we fill the joint with more viscous silicone oil of 100 cSt during the experiment in a normal environment. Therefore, with the NTP (normal temperature and pressure), we could simulate the effect of the pressurized oil on the underwater joint owing to the similar loss torques. It considerably simplifies the experiment.

\begin{figure}[!h]
   \centering
\includegraphics[width=.5\textwidth]{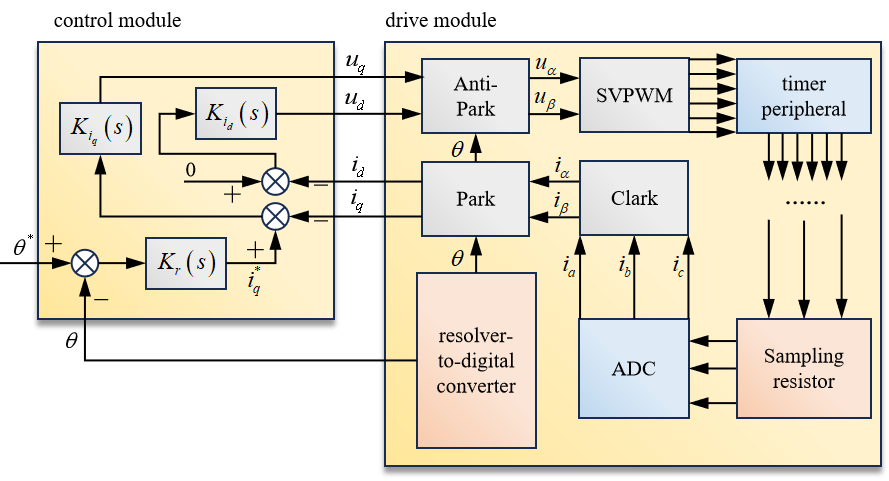}
   \caption{Control framework including current controllers.}
   \label{fig_controlf}
\end{figure}

Generally, the inputs of the motor drive module are the decoupled torque voltage $u_q$ and the magnetic flux voltage $u_d$. As mentioned in Section \ref{sec4}, the control module outputs a decoupled torque current $i_q$ to actuate the joint system. To ensure an accurate and rapid tracking of $i_q$, the robust postion controller is follwed by a PID current controller, which outputs the drive input $u_q$. Another PID controller is designed to ensure $u_d = 0$ for concentration on motor torque control. The setting time of the current response is 5 ms or so. Thus, we consider the current controllers in the closed-loop system as a simple time delay. Fig. \ref{fig_controlf} illustrates the control framework including the current controllers.

\begin{figure}[!h]
\centering
\subfloat[]{\includegraphics[width=.32\textwidth]{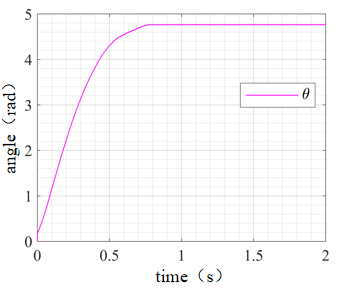}}
\hfil
\subfloat[]{\includegraphics[width=.32\textwidth]{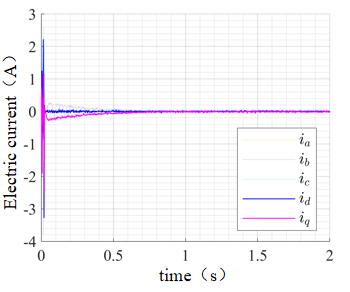}}
\hfil
\caption{Step responses during the position tracking experiments. (a) Position response. (b) Torque current response.}
\label{fig_pos_step_response}
\end{figure}

In the position control experiment, a reference position $\theta^{*}=3\pi/2$ is transmitted to the motor control board. Subsequently, this board implements the relevant algorithms and outputs PWM (pulse width modulation) signals to the concatenated gate driver and three-phase inverter to actuate the motor of the joint. The position of the joint output shaft is measured by the resolver and fed back to related algorithm from the resolver decoder board to the motor control board via CAN bus. Fig. \ref{fig_pos_step_response} shows this transient step response. The measured joint position tracks the reference signal in a settling time of 0.76 s without overshooting. Besides, the corresponding torque current $i_q$ is also within the acceptable range. It validates the feasibility of our control framework.

\section{Conclusions}\label{sec6}
For the development of an electric oil-filled joint system working in an underwater environment, this study integrates the mechanical, drive, and control modules of the joint. To sustain the harsh ambient pressure, a compensation technique is implemented and the relevant dynamics is analyzed. A joint structure optimization framework is proposed to boost the sealing performance, where the filled hydraulic oil is prevented from leakage. To consider the parametric perturbations and unknown disturbances caused by the viscous oil, an uncertain dynamic model of the joint is constructed and a robust position controller is designed using $\mu$-synthesis. We carry out relevant experiments, which denote that the joint system could achieve the demand for position control under the parametric uncertainty and unknown disturbance.

Despite the achievements, further improvements remain to be made. Till now, we only focus on one single joint. The complete underwater manipulator composed of multiple joints and the relevant underwater vehicle manipulator system (UVMS) need further research. Besides, we will also study data-driven methods to express the dynamics of the oil-filled joint system. In the future, experiments in the real ocean environment should be conducted.

\printcredits

\bibliographystyle{cas-model2-names}

\bibliography{cas-refs}






\end{document}